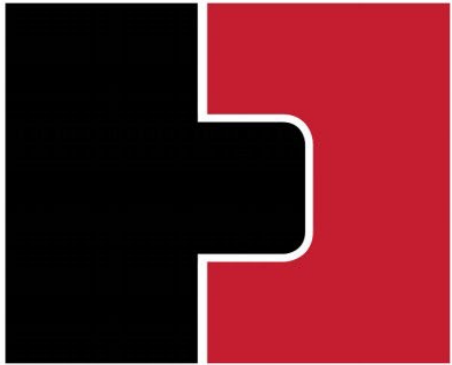

# SmartShuttle: Model Based Design and Evaluation of Automated On-Demand Shuttles for Solving the First-Mile and Last-Mile Problem in a Smart City

Authors: Sukru Yaren Gelbal, Bilin Aksun-Guvenc, Levent Guvenc

# FINAL RESEARCH REPORT

Contract No. DTRT-13-GUTC-26

## Introduction

The final project report for the SmartShuttle sub-project of the Ohio State University is presented in this report. This has been a two year project where the unified, scalable and replicable automated driving architecture introduced by the Automated Driving Lab of the Ohio State University has been further developed, replicated in different vehicles and scaled between different vehicle sizes. A limited scale demonstration was also conducted during the first year of the project. The architecture used was further developed in the second project year including parameter space based low level controller design, perception methods and data collection. Perception sensor and other relevant vehicle data were collected in the second project year. Our approach changed to using soft AVs in a hardware-in-the-loop simulation environment for proof-of-concept testing. Our second year work also had a change of localization from GPS and lidar based SLAM to GPS and map matching using a previously constructed lidar map in a geo-fenced area. An example lidar map was also created. Perception sensor and other collected data and an example lidar map are shared as datasets as further outcomes of the project.

## Brief Problem Description

A major component of mobility in a smart city is the use of fully electric driverless vehicles that are used for solving the first-mile and last-mile problem, for reducing traffic congestion in downtown areas and for improving safety and helping in the overall reduction of mobility related undesired emissions. Currently available Smart Shuttle solutions have serious interoperability problems due to the low volumes of production and due to the fact that they are developed and manufactured by small startup companies in contrast to OEMs with their series production capability and large R&D departments. Current Smart Shuttle sensing and automation architectures are, therefore, also not easily scalable and replicable. Success of Smart Shuttles in Smart Cities requires an interoperable, scalable and replicable approach which is what this project addresses through model based design techniques.

## Project Approach, Methodology and Results

Our model based design approach uses a unified software, hardware, control and decision making architecture for low speed smart shuttles that is scalable and replicable. Robust parameter space based design is used for easily scalable low level control systems development. Our model based design approach uses model-in-the-loop and hardware-in-the-loop simulations before road testing. This method was demonstrated in proof-of-concept testing/deployment in non-public areas including parking lots.

A unified scalable and replicable architecture and the hardware-in-the-loop simulator for automated driving were prepared in this project. Extensive model-in-the-loop and hardware-in-the-loop simulations were used for testing the automated driving system in the lab setting first. Testing included communication with other vehicles and instrumented traffic lights using a DSRC on-board unit (OBU) modem and a DSRC road-side unit (RSU) modem that were added to our connected and autonomous driving hardware-in-the-loop simulator. Four different platforms in two different vehicle size categories were used in experimental work for replicability and scalability. These vehicles were used in generating the experimental results some of which are given in this report. Please refer to the enclosed papers for more detailed results.

Our work on unifying the structure with scalability and replicability is to create a standard base for hardware structure along with a library to be used by developers for faster and easier automation of vehicles. The hardware structure includes different types of sensors to achieve enough coverage, resolution and also robustness to external disturbances. Data from these sensors is processed by a high processing power computer to create meaningful information, which is used by a low-level controller, i.e. a dSpace MicroAutobox in our vehicles, to drive the vehicle autonomously by interfacing with actuators and sending necessary commands. The unified architecture is shown in Figure 1.

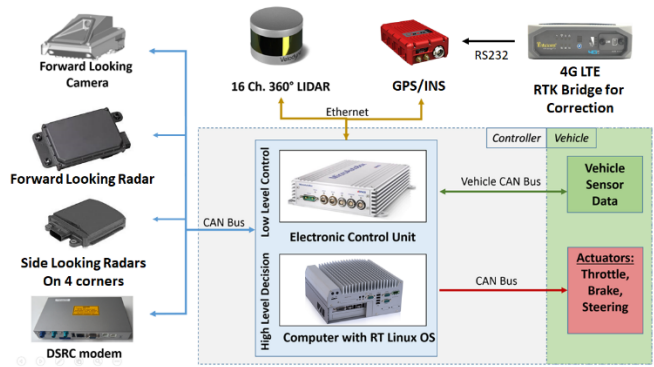

Figure 1. Unified architecture.

## *Replicability and Scalability*

Using this unified architecture, two different sized vehicles were automated. These are a sedan, representative of passenger vehicles and a small neighborhood electric vehicle, representative of small on-demand electric shuttles. Perception sensors such as Lidar, Camera, Radars were implemented as well as a GPS Sensor with RTK correction for localization. The dSpace MicroAutobox unit is used for low level controls and an in-vehicle Linux PC with a GPU is used for sensor data computation. Moreover, DSRC (dedicated short-range communications) radios are added to have the capability of communicating with other vehicles, pedestrians, bicyclists and infrastructure. Pictures of the vehicles and the implemented hardware as well as the evaluation of the scalability approach is presented in our papers and is shown graphically in Figure 2 here. On the top are the two vehicles we automated first. These are a 2015 Ford Fusion and a Dash EV neighborhood electric vehicle. Both vehicles were made drive-by-wire first and the sensor, hardware and computing architecture of the Ford Fusion sedan was scaled down to the Dash EV vehicle. The low level longitudinal and lateral controllers developed for the Ford Fusion using the parameter space design approach were also scaled down and used in the Dash EV after some re-tuning due to changes in vehicle parameters. The architecture/results were, then, replicated as we migrated from the

2015 Ford Fusion to the 2017 Ford Fusion and from the white Dash EV to the red Dash EV both in the bottom of Figure 2. The drive-by-wire system for the Ford Fusion vehicles uses a commercially available system that uses CAN bus commands to control the actuators. This drive-by-wire was moved from the 2015 Ford Fusion to the 2017 Ford Fusion with a software upgrade that enabled shift-by-wire on top of the previously available throttle, brake and steer-by-wire. The drive-by-wire systems of both Ford Fusion vehicles were based on the original OEM actuators such that when the drive-by-wire system was turned on, we had the original vehicle which we can legally drive on public highways. The drive-by-wire system of the Dash EV vehicle was prepared in-house and is illustrated in Figure 3 for our first Dash EV vehicle. The throttle in this electric vehicle uses a potentiometer which was changed according to the autonomous driving controller using an additional circuit that changes the throttle potentiometer reading. A linear actuator was used to pull/push the brake pedal for brake-by-wire and a smartmotor and an extra half steering linkage was added to provide driver independent steering as needed for steer-by-wire operation. An extra electronic circuit with a manual/autonomous conversion switch was added for switching between normal vehicle operation and drive-by-wire operation. As the two Dash EV vehicle dimensions and characteristics were very similar, the drive-by-wire system was easily transferred from the first vehicle (white one on top in Figure 2) to the new one (red one in bottom of Figure 2). A remote e-stop kill switch with an RF link is also being added to the new Dash EV vehicle. Our new Dash EV vehicle has a license plate like or Ford Fusion vehicle and can be driven on public roads in the manual mode.

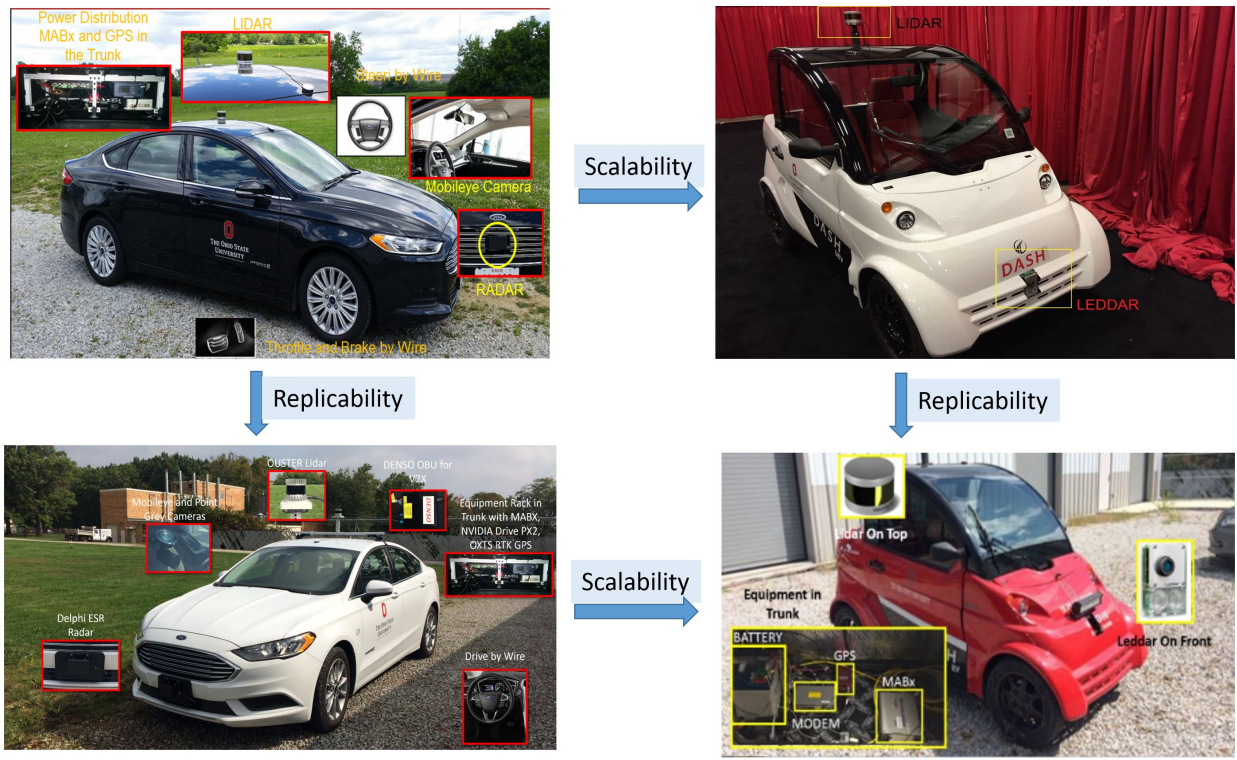

Figure 2. Scalability and replicability.

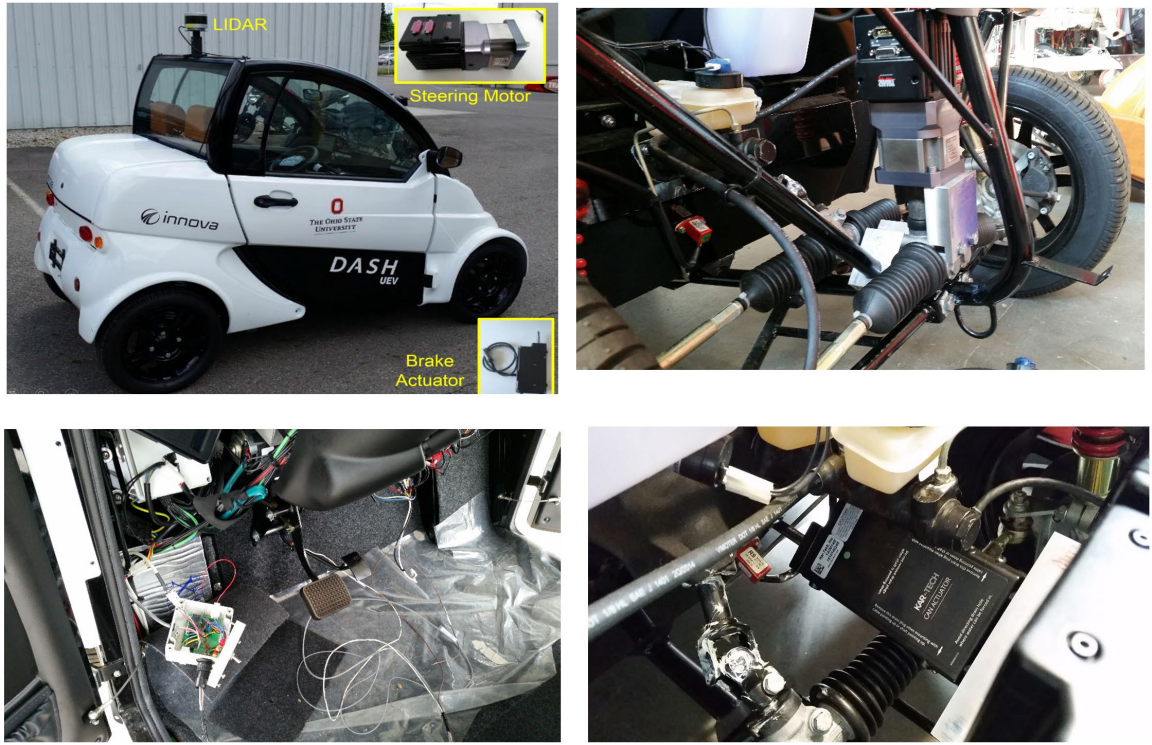

Figure 3. Dash EV drive-by-wire system.

Along with the unified architecture, a unified Simulink library was also created. This library shown in Figure 4 consists of different types of blocks, including low-level control blocks for steering, throttle, brake, shift; sensor blocks for receiving data from the sensors in order to have environment perception and localization; and finally control and decision-making blocks for low and high level control of the autonomous vehicle. We are currently extending this library for use with NVIDIA Drive PX 2 GPUs. It is slightly modified for CarSim soft sensors and then used in the hardware-in-the-loop (HIL) simulations presented later in the report.

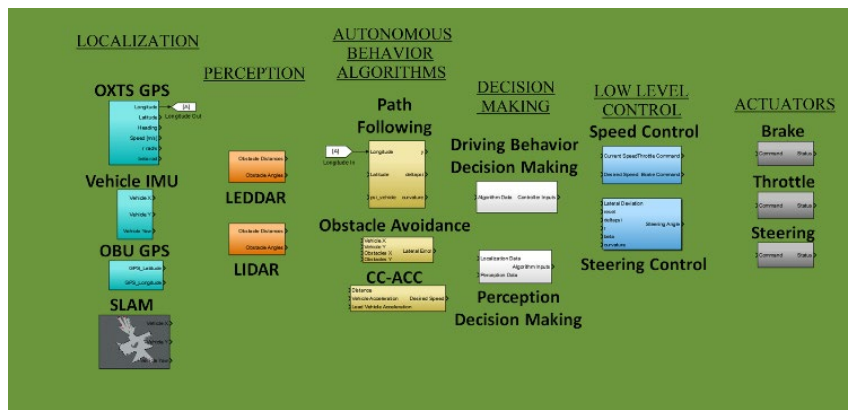

Figure 4. Simulink vehicle automation library.

## Controls - Low Level Longitudinal Control

Parametric modeling is the first step necessary in designing controllers. For the longitudinal speed control of the Ford Fusion sedan, we performed system identification on experimental data of different throttle step input and obtained a simple first order model. A typical example is presented here for a 15% throttle step input. The experimental data is recorded from the vehicle by giving a step input to the throttle and then recording the vehicle velocity [1]. A first order transfer function was then fitted to the vehicle velocity data. The transfer function obtained through this method for 15% throttle input is given in Equation 1.

$$G(s) = \frac{0.1515}{s + 0.07496} \tag{1}$$

The curve fit comparison of experimental, Simulink model and constructed Carsim model are shown in Figure 5 for 15% step throttle input. The procedure is repeated for different throttle openings to obtain a family of plants which are then used in a parameter space design to prepare a scheduled PID longitudinal speed controller.

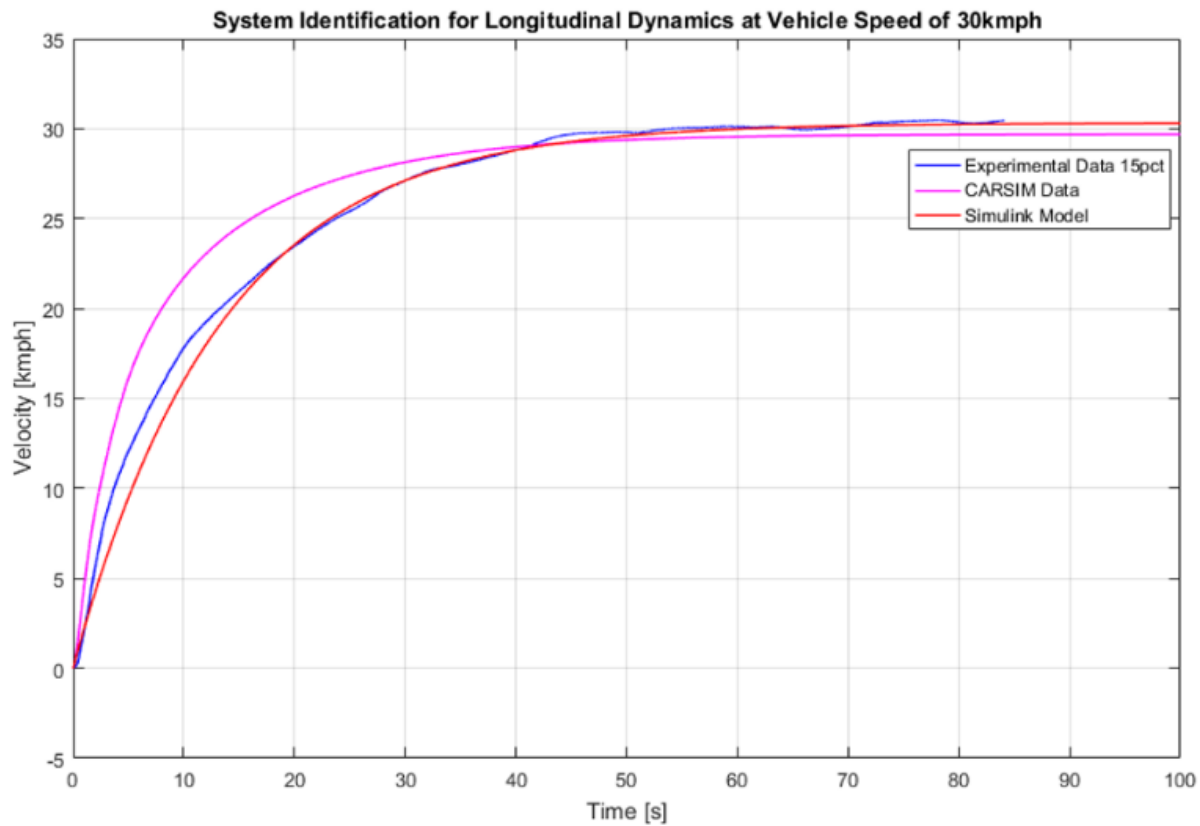

Figure 5. Velocity graph for experimental result, Carsim model and Simulink model for 15% throttle.

It should be noted that the longitudinal dynamics of the vehicle is highly non-linear and changes with throttle position. While a simple curve fit at one throttle opening is shown and used for initial step, a more comprehensive multi-model determination based on the experimental data we collected is in progress and will be used in our future work.

### *Controls - Low Level Lateral Control*

Lateral controller design also requires the development of a model of the lateral dynamics of the vehicle. Both Simulink and Carsim models were developed for both the sedan and neighborhood electric vehicles. The model building process involved the use of testing machines (for the sedan) as shown in Figure 6 for our 2015 Ford Fusion sedan and identification of parameters using standard tests (for the sedan and the neighborhood electric vehicle). The same models and low level controllers designed based on them as the architecture and controllers were replicated and transferred from our 2015 to 2017 sedan. A similar replication of model and low level controllers was used without problems as we migrated from one Dash EV vehicle to the next one.

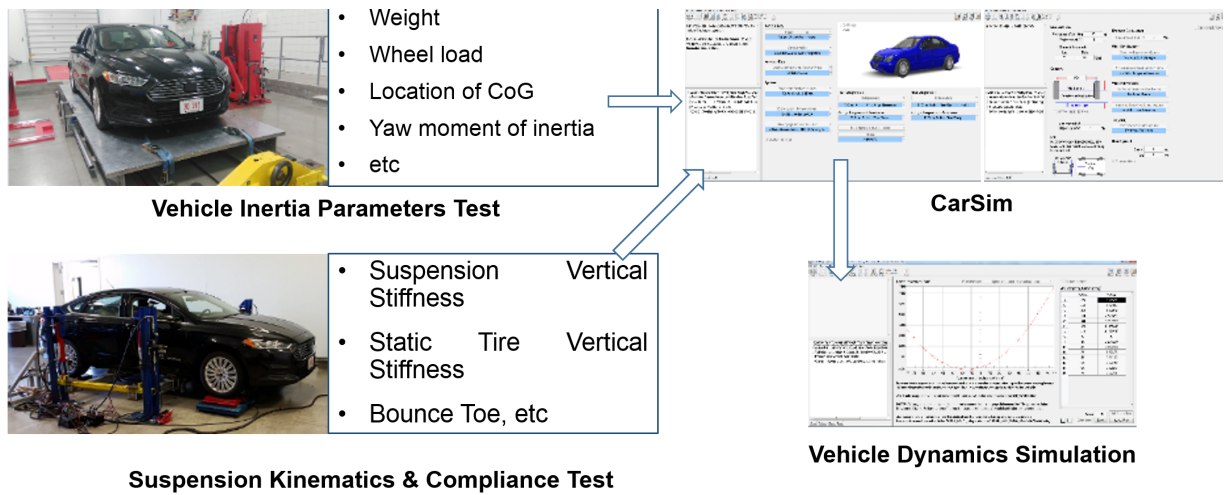

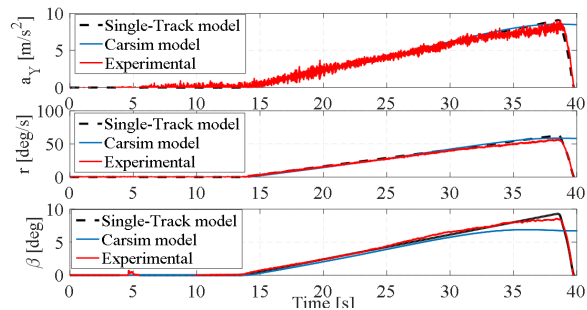

Figure 6. Measurement of vehicle model parameters of 2015 Ford Fusion.

Measured, identified and estimated parameters for the sedan and small electric vehicles are listed in Table 1.

Table 1. Vehicle lateral model parameters.

|     | Ford Fusion | Dash |
| --- | --- | --- |
| m | 1977.6 kg | 350 kg |
| J | 3728 kgm$^2$ | 350 kgm$^2$ |
| lf | 1.3008 m | 1.06 m |
| lr | 1.54527 m | 0.96 m |
| R | 0.3225 m | 0.24 m |
| Cf | 1.9e5 N/rad | 1.8917e4 N/rad |
| Cr | 5e5 N/rad | 1.8917e4 N/rad |
| $k_p$ | 0.15 | 0.9272 |
| $k_d$ | 0.1 | 0.0801 |

Satisfactory lateral control requires high accuracy of path tracking, and robustness to system parameter variations like vehicle load, speed and tire cornering stiffness. Figure 7 shows variation regions of these parameters for our two vehicle platforms, the low-speed shuttle Dash and the Ford Fusion sedan. The design procedure for lateral control we used satisfies D-stability and robust performance and is easily replicable and scalable to other vehicle platforms.

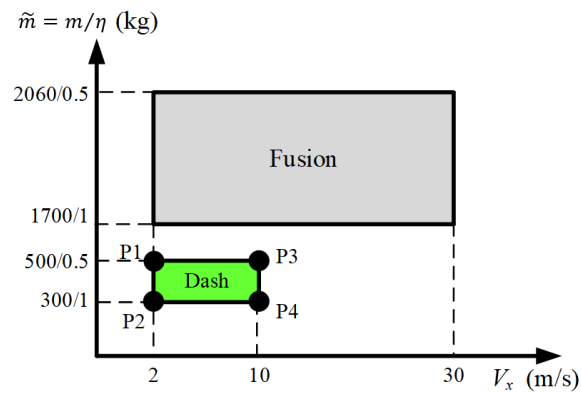

Figure 7. Uncertainty region of vehicle mass $m$, longitudinal speed $V_x$ and parameter $\eta$ for Dash and Fusion experiment platforms.

The system roots are confined in the D-stability region (Figure 8) in the complex plane to satisfy requirements like settling time, damping ratio and bandwidth. The mixed sensitivity criterion is also raised to ensure robust performance in frequency domain. The boundaries of the D-stability region, along with the points satisfying the mixed sensitivity critical criterion, are mapped to the parameter space of control parameters, $k_p$ and $k_d$, for the robust proportional-derivative (PD) controller. Figure 9 shows an example of the control parameter space at one operating condition. The PD control parameters are chosen from the selectable region after overlapping parameter space for all operating conditions. A model regulator is added in combination with the robust PD controller to further reduce tracking error (Figure 10). The model regulator works by rejecting the road curvature as a disturbance for improving path tracking error in the presence of unstructured and parametric error in the vehicle model. Sometimes, we use a classical controller plus a feedforward controller since we usually have preview of the road ahead. This approach also improves path tracking performance considerably.

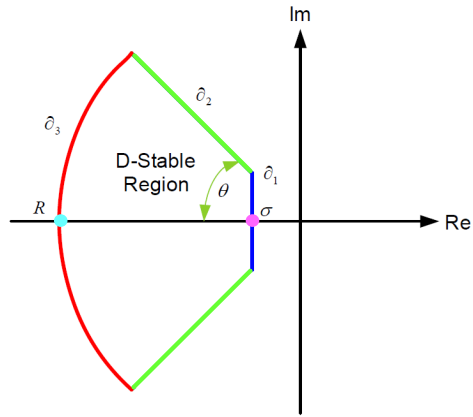

Figure 8. Illustration of D-stability region in the complex plane.

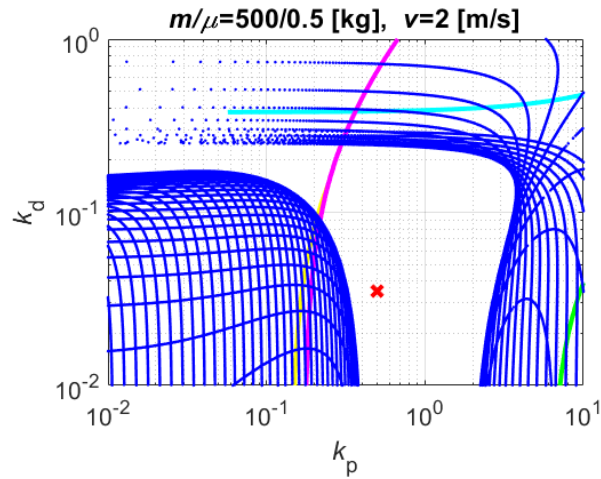

Figure 9. Parameter space of $k_p$ and $k_d$ at one uncertainty vertex

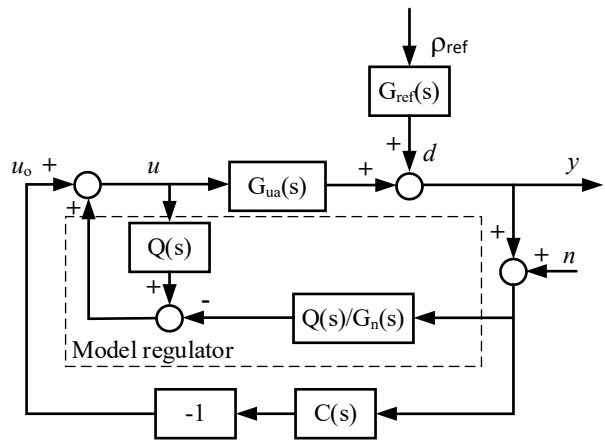

Figure 10. System diagram with the PD controller and model regulator.

## Localization - eHorizon

eHorizon is an electronic horizon equipment with a detailed map inside the device that can extract road information such as GPS, speed limits, heading direction, locations of the intersections, road curvatures, traffic light and STOP sign positions. In short, it gives us a 1-2 km preview of the map ahead. Due to the information it can provide, eHorizon is a powerful tool for Path Planning and Eco-Route Planning applications, as well as Fuel Economy studies for Automated and Connected Vehicles. eHorizon uses ADASIS v2 (ADAS Interface Specification) protocol and sends map related information over a CAN network. Map data is transmitted, deconstructed by eHorizon, transmitted in a series of messages, and then reconstructed by the device that eHorizon is connected to, such as the MicroAutoBox (MABX).

The ADASIS protocol is founded on the idea of a "horizon". The horizon is a defined distance ahead of the vehicle in which relevant map features are provided. The horizon data mainly comprises of paths (possible routes that the vehicle could take) and profiles (data about those routes such as traffic signals and slope), as seen in Figure 11. eHorizon calculates the Most Probable Path (MPP) that the vehicle is expected to follow and sends the information and profiles on this main path to the user.

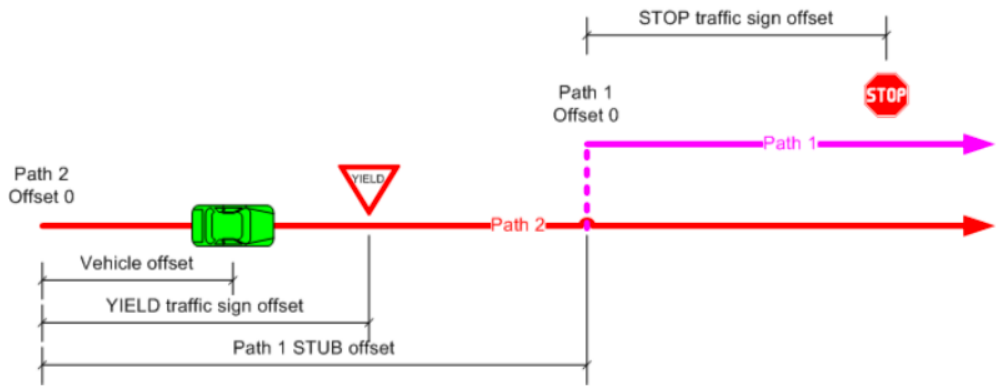

Figure 11: eHorizon Paths and Profiles

The eHorizon unit comes with its own software for visualization of the route the vehicle is following. An example of how the eHorizon software looks during travel can be seen in Figure 12. The orange arrow in Figure 12 indicates the heading direction and position of the vehicle. The solid blue line on the map illustrates the Most Probable Path the vehicle is expected to follow. The green circles in the left side of Figure represent the traffic lights and the red squares are used to illustrate the STOP sign locations.

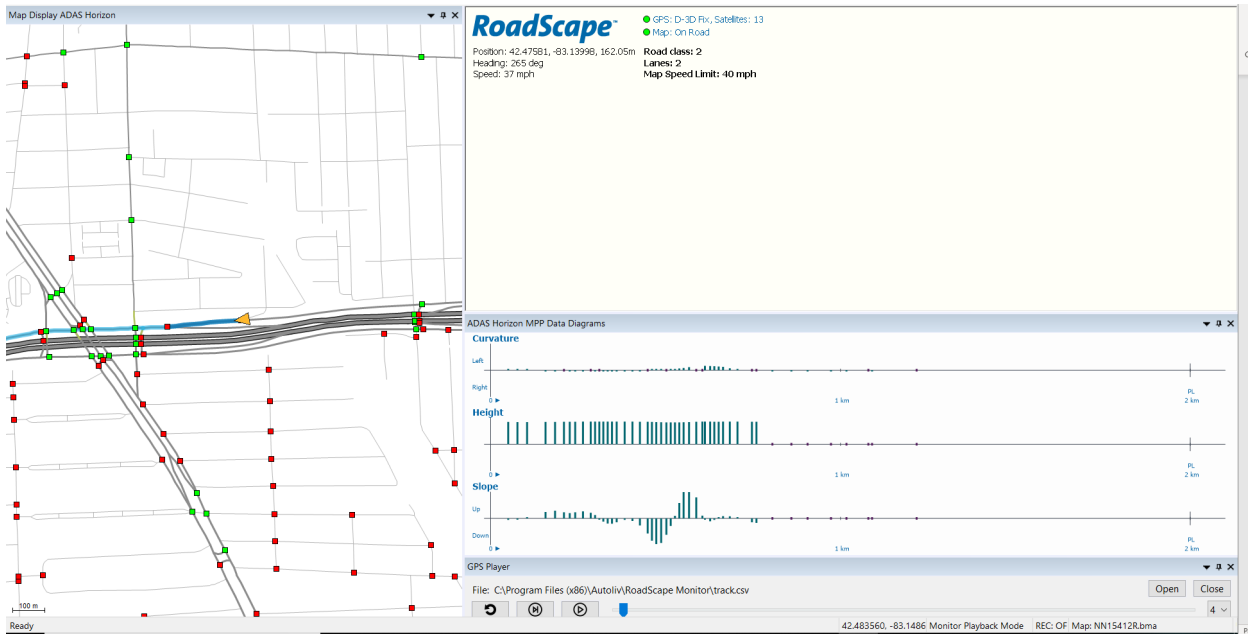

Figure 12: eHorizon window during in-vehicle experiment

As an example, some experiments were run with the eHorizon Autoliv RoadScape unit of the Automated Driving Lab (ADL) which was placed in the Ford Fusion sedan. For a drive starting from and ending at our lab building at CAR-West, some results are summarized in Figure 14. The 1$^{st}$ subplot in Figure 13 shows the GPS points acquired with the eHorizon unit during the actual testing. Another information that could be extracted from eHorizon was the heading angle, and the results were plotted in the 2$^{nd}$ subplot. eHorizon was able to provide what type of road the vehicle was on, and these results were plotted in the 3$^{rd}$ subplot. Finally, the speed limit of the route the vehicle travelled on was gathered from eHorizon and was plotted in the 4$^{th}$ subplot.

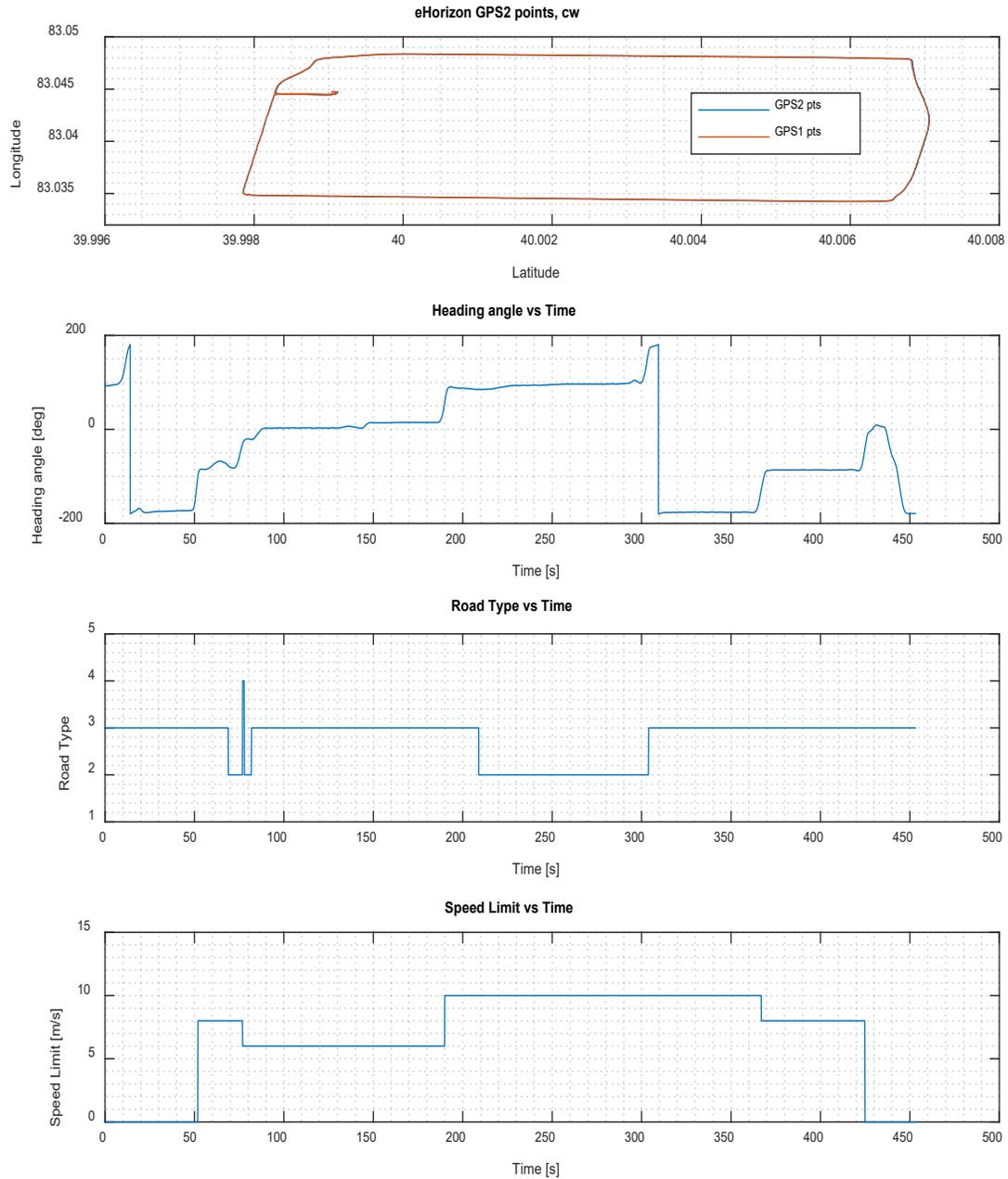

Figure 13: eHorizon results

The eHorizon unit is also able to provide upcoming traffic light and STOP sign information. To get the upcoming traffic sign information, an in-vehicle experiment was conducted with the eHorizon unit in Columbus and the test route can be seen in Figure 14. In Figure 14, red squares represent the position of the STOP signs and the yellow circle represents the position of the traffic light. Since we believe that the map is just like another sensor, the e-horizon or map preview sensor is also an important part of our unified autonomous driving architecture.

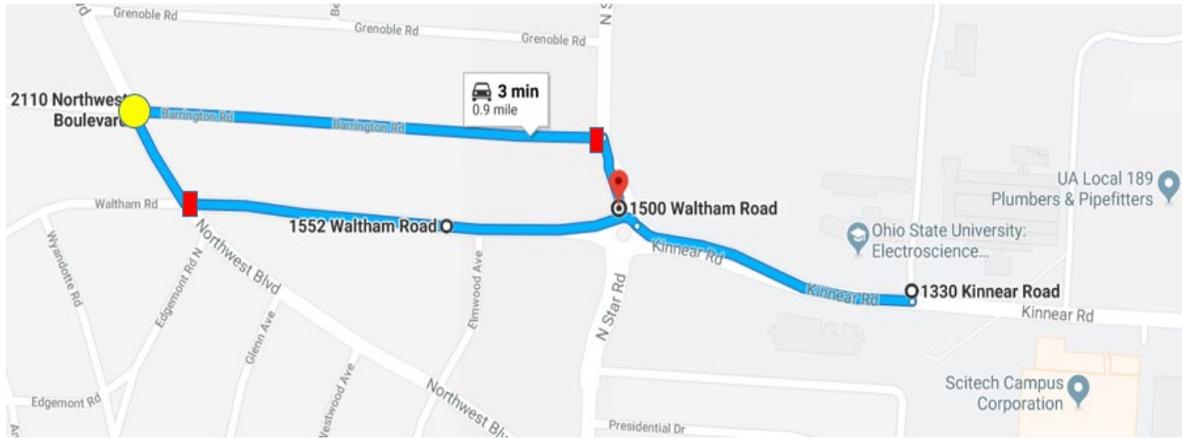

Figure 14: Test Route for eHorizon test

On the test route, shown in Figure 14 traffic sign information was gathered and can be seen below in Figure 15. Whenever the upcoming traffic sign was a STOP sign, an internal variable in the eHorizon unit provided value 2. Similarly, whenever the upcoming traffic sign was a traffic light, the same internal variable in the eHorizon unit provided the value 1. Even though there were a total of 3 traffic signs for this route, the eHorizon unit changed between 1 and 2 more than 3 times. The reason for that is because the eHorizon unit predicts the Most Likely Path the vehicle is going to travel and depending on that path, shows the upcoming traffic sign info. If the driver does not follow the Most Probable Path predicted by eHorizon, then eHorizon collects the correct traffic sign information for the route the vehicle is actually following from the map.

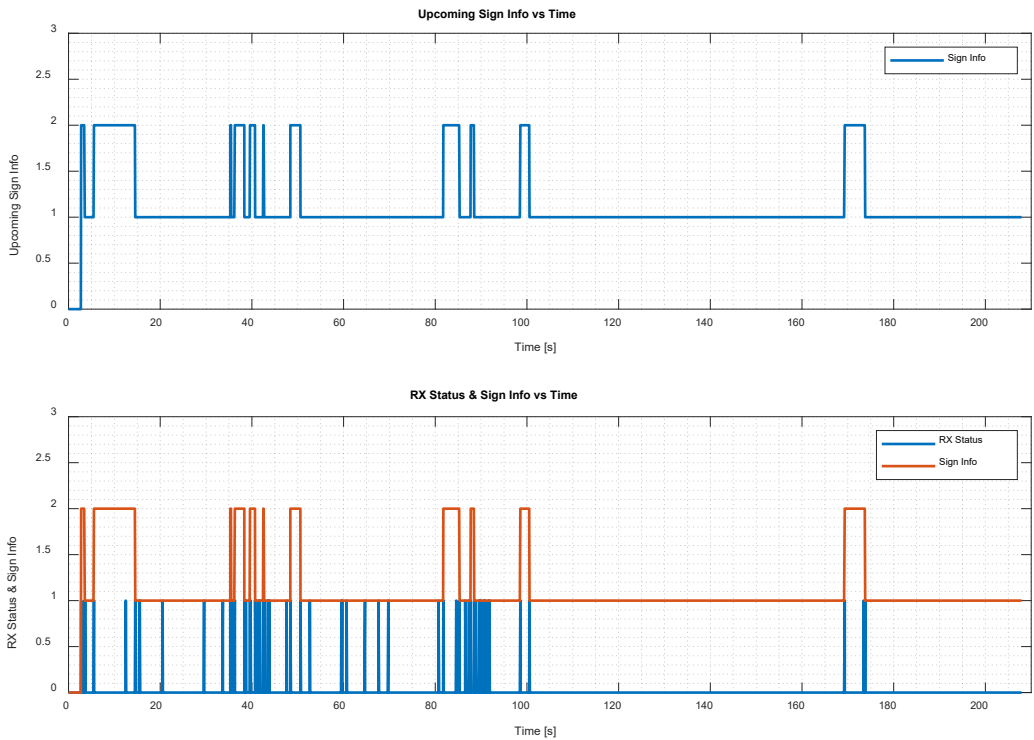

Figure 15: eHorizon providing upcoming traffic sign info

*Map Building/Matching Lidar Maps*

In the research and application of autonomous vehicle, to know about the current state and position of the ego vehicle is very essential. It is a demanding task to localize the ego vehicle and estimate the state of the ego vehicle as this information is needed by other functions in autonomous driving. To estimate the vehicle state, odometry and IMU data are used in different autonomous driving research and application. Also, in the past few decades, the development high accuracy dual antenna GPS encourages good performance of vehicle localization especially when real time kinematic corrections are used. In the meanwhile, the navigation suite is able to provide elaborate information about vehicle state alongside with global localization. However, high accuracy GPS cannot work well when the signal receiving gets weak due to the blockage from the environment, such as bridges and skyscrapers in an urban area. Thus, localization methods based on camera and Lidar sensors are developed.

In our study, to ensure the precision of vehicle localization and vehicle state estimation, the GPS localization method with OXTS Navigation suite and map matching based localization are implemented.

*RTK-GPS*

For the GPS based localization, the RTK-GPS combination are used. The OXTS GPS navigation suite provides location information that is up to 20 cm accuracy with IMU data and has a built in Kalman filter. The RTK bridge from Intuicom® is a real time kinetic localization equipment that receives location information accessing the ODOT VRS System provided by Ohio Department of Transportation [2]. Combining the two devices under strong signal receiving, the accuracy of global localization is as high as 2 cm. Typical accuracy values were around 5 cm while this dropped to 20 cm in downtown areas where some GPS satellite signals were lost.

*Map Building*

Map matching based localization technique is based on algorithms of point cloud transformation and matching. The requirements of applying map matching in localization include well built map and good matching algorithm. In this project, a 3D point cloud map was built for map matching based localization using 3D lidar data.

The 3D map is built from 3D point cloud data captured by a Lidar sensor. And the mapping technique is based on normal distribution transform [3] for scan matching and ndt_mapping [4], that is, align different frames of point cloud data by transforming them into same reference frame and overlap the parts where they have similar point cloud.

For the normal distribution transform of a single frame point cloud data, a piecewise-continuous probability density function is generated as in Figure 16. Mean and covariance matrices are calculated from the point cloud within the grid. The higher the probability of point cloud lying in a grid, the larger is the value of its probability density.

Thus, using the probability density function gives a good description of Lidar point cloud with respect to the density and probability of showing up.

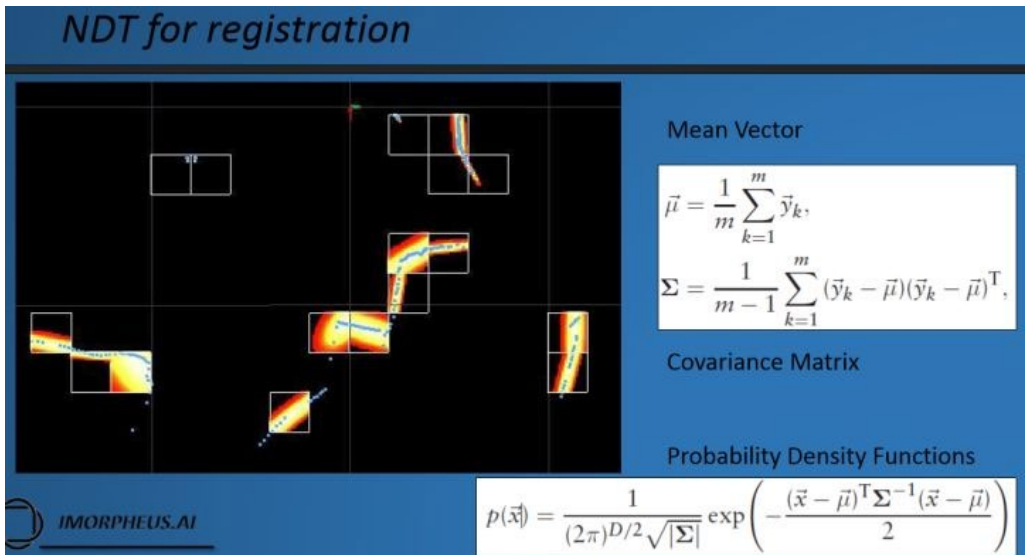

(a)

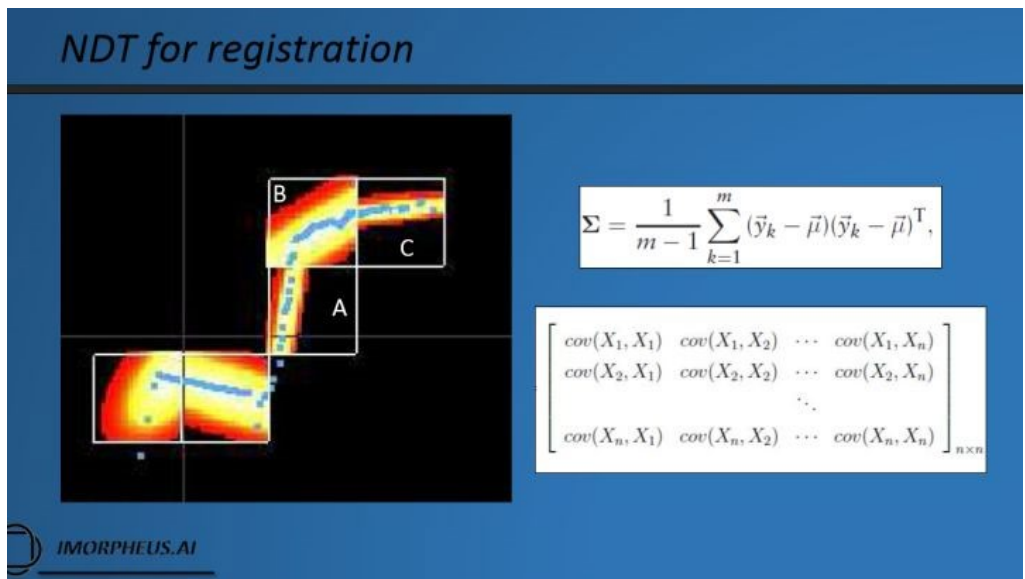

(b)

Figure 16. NDT for 2D laser scan data. Blue dots are original point cloud, probability distribution is shown as red and yellow area. (a)The black area above is the laser scan to be transformed. Grid the laser scan and generate a density function of each grid. (b) Difference in probability of different grids. (IMORPHEUS.AI)

For 3D point cloud data, the covariance matrix is composed from the eigen value and eigen vectors of the 3D point cloud, but the idea is the same with 2D point cloud.

## Map Matching based Localization

Given a 3D point cloud map, the map matching based localization is used here as a compensation for GPS localization. Recently, various algorithms of scan matching are developed which are able to find the transform between two point clouds. Iterative Close Point (ICP) algorithms [5] and Normal Distribution Transform (NDT) are two commonly used algorithms with good performance. The approach we use in this study is the NDT method along with odometry extrapolation. [6-7].

For the NDT algorithm, the inputs are pre-built 3D point cloud map and the current point cloud scanned by Lidar Sensor. By matching those two point cloud data, a transformation will be generated as $T = (t, R)$ where $t$ is the translation of current point cloud scan in the map and $R$ is the rotation. From the transformation information, the goal of localization in the reference frame of map is achieved. Figure 17 shows a graphical illustration of the NDT algorithm. With odometry extrapolation, a better guess of initial transformation for matching is acquired so that the precision of matching between current scan and pre-built map is increased.

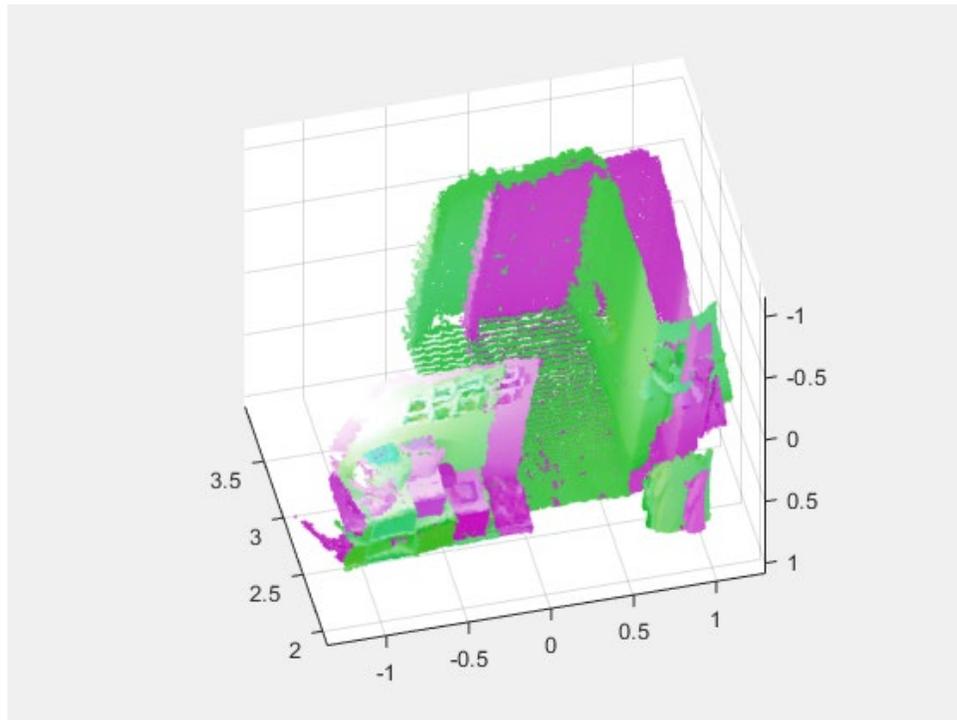

Figure 17. Sketch of NDT scan matching. The green point cloud and pink point cloud are aligning in the same reference frame after transformation.

## Example Lidar Maps

Using the methodology discussed in previous part, we have built the map for several places in Columbus, Ohio. Examples are shown in Figures 18-20. Figure 18 shows the Scioto Mile route in downtown Columbus where an AV shuttle by a commercial AV tech company is operating seven days a week. Figure 19 shows the main route of the planned Ohio State University pilot AV shuttle deployment. The route in Figure 20 is the route we use in our initial tests of localization and autonomous driving. We have used these maps for map matching based localization and compared the results with high accuracy GPS recording of the same experiment and determined that accurate localization was achieved in real time operation.

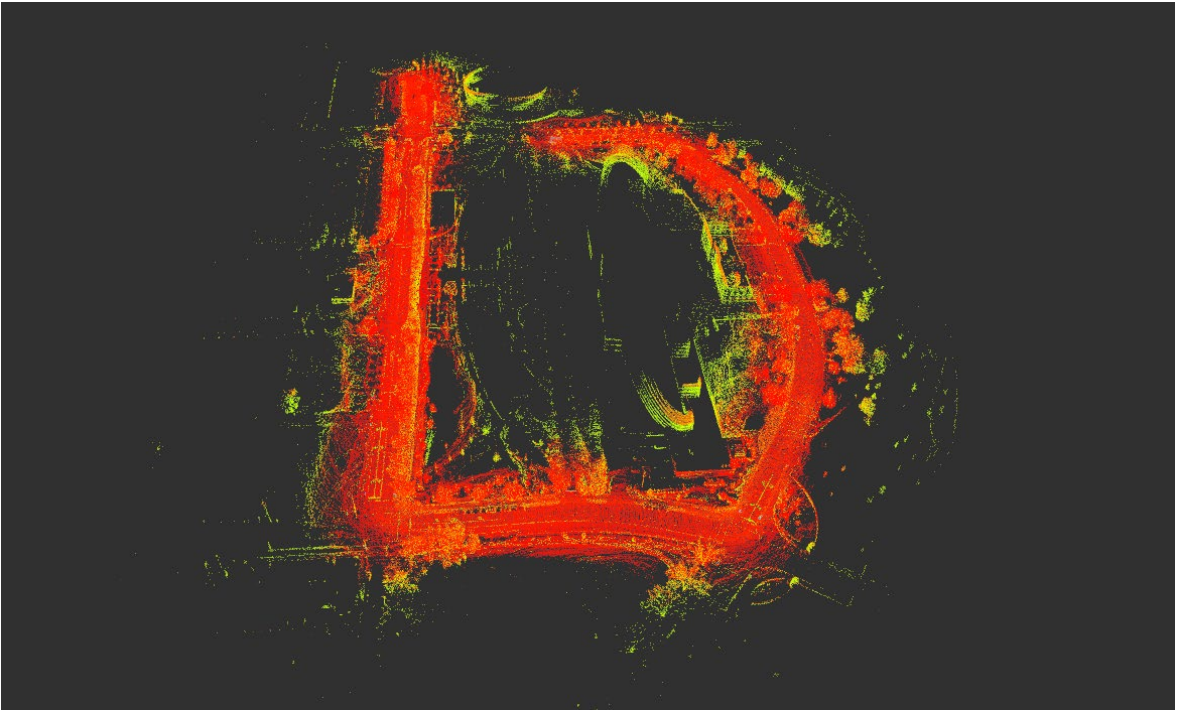

Figure 18. 3D Point cloud Map of Scioto Mile around COSI downtown Columbus.

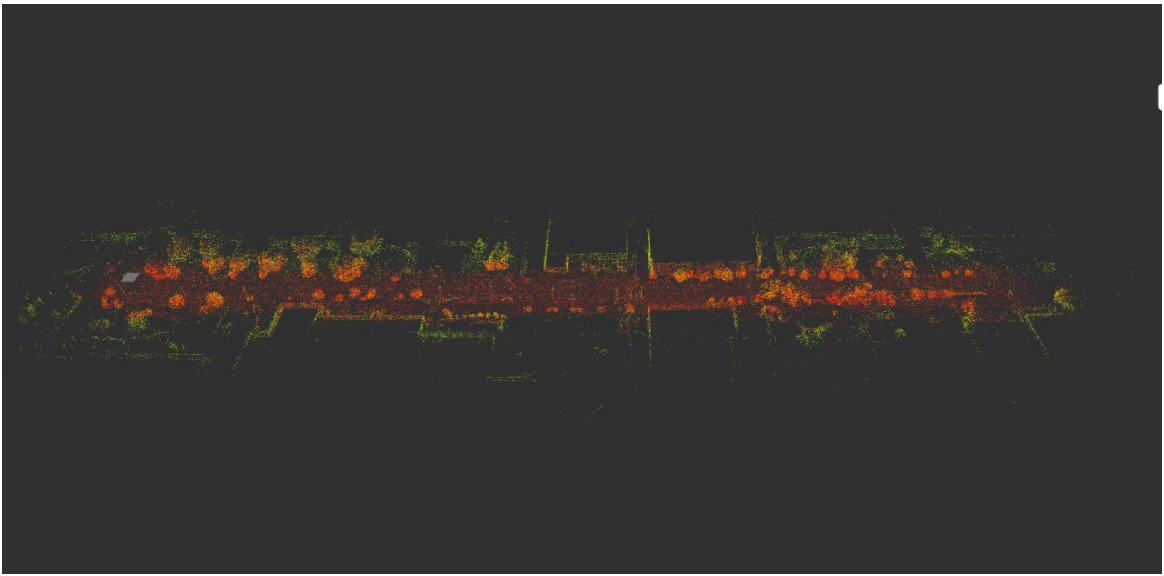

Figure 19. 3D Point cloud Map of OSU AV shuttle pilot route from CAR to CAR West.

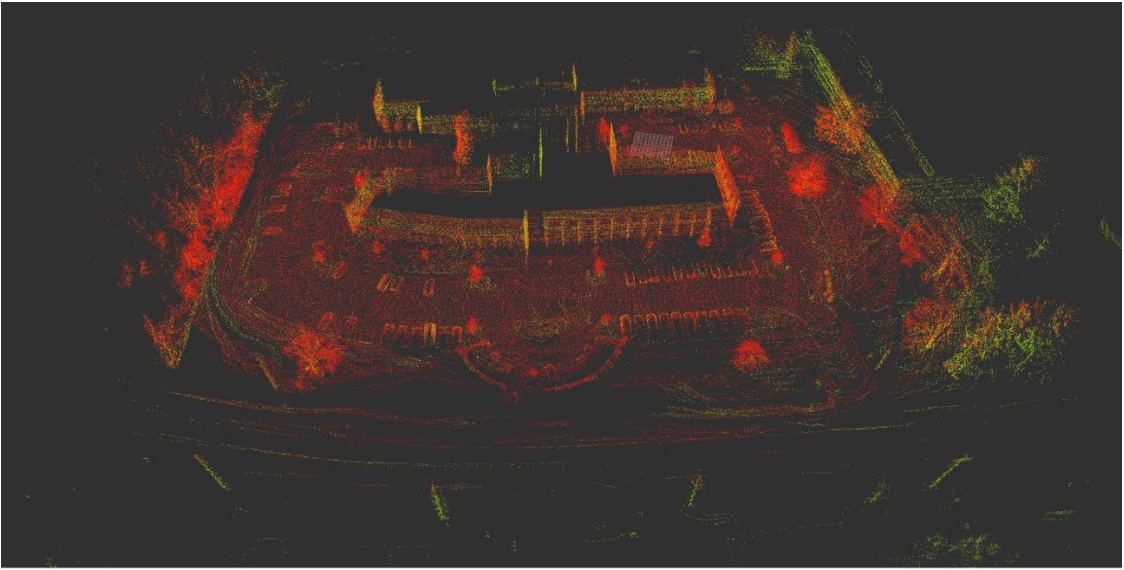
Figure 20. 3D Point cloud Map.of parking lot route at CAR West.

In this part of the report, different techniques for vehicle localization were presented and applied to real time the autonomous vehicle operation. We emphasized map matching based localization in comparison to RTK GPS localization. NDT based map building and map matching were implemented for building 3D point cloud map and solving the problem of vehicle localization and state estimation.

## *Path Tracking*

The low level automated driving tasks are lateral and longitudinal control. The path determination and path tracking error computation are described briefly in this section. The path tracking model consists of two parts, which are offline generation of the path and online calculation of the error according to the generated path. These parts are explained in following subsections.

**Offline Path Generation**

The path following algorithm employs a pre-determined path to be provided to the autonomous vehicle to follow. This map is generated from GPS waypoints where these points can be pulled from an online map or can be collected through recording during a priori manual driving. These data points are then divided into smaller groups named segments with equal number of data points for ease of formulation. These segments are both used for curve fitting and velocity profiling through the route. After dividing the road into segments, a process of fitting a third order (or higher order) polynomial is performed as

$$X_i(\lambda) = a_{xi}\lambda^3 + b_{xi}\lambda^2 + c_{xi}\lambda + d_{xi}$$
$$Y_i(\lambda) = a_{yi}\lambda^3 + b_{yi}\lambda^2 + c_{yi}\lambda + d_{yi}$$

(2)

where *i* represents the segment number and the terms *a, b, c, d* are polynomial fit coefficients for the corresponding segment. Fitting the data points provides effective replication of the curvature that the road carries and also eliminates the noise in the GPS data points. To provide a smooth transition from one segment to another by satisfying continuity of the polynomials and their first derivatives in X and Y, we use

$$X_i(1) = X_{i+1}(0)$$
$$Y_i(1) = Y_{i+1}(0) \tag{3}$$

The X and the Y points derived from the GPS latitude and longitude data using a degree to meter conversion, are fit using a single parameter λ, where λ is the variable for the fit which varies across each segment between *0* to *1*, resulting in

$$\frac{dX_i(1)}{d\lambda} = \frac{dX_{i+1}(0)}{d\lambda} \tag{4}$$

$$\frac{dY_i(1)}{d\lambda} = \frac{dY_{i+1}(0)}{d\lambda} \tag{5}$$

**Error Calculation**

After the generation of path coefficients, an error is calculated for the lateral controller to use as input. Heading and position of the vehicle is provided by means of localization, in this case either SLAM, map matching or GPS. Using these, the location of the car with respect to the path, in other words, the deviation from the path is calculated. This approach reduces both oscillations and steady state lateral deviation compared to calculation with respect to position only. In order to find an equivalent distance parameter to add to the first component distance error, a preview distance $l_s$ is defined. Then, the error becomes,

$$y = h + l_s \tan(\Delta\psi) \tag{6}$$

where Δψ is the net angular difference of heading of the vehicle from the heading tangent to the desired path and *y* is the total error of the vehicle computed at preview distance $l_s$ as is illustrated in Figure 21. Finally, this error is fed to a robust controller which controls the actuation of steering of the vehicle.

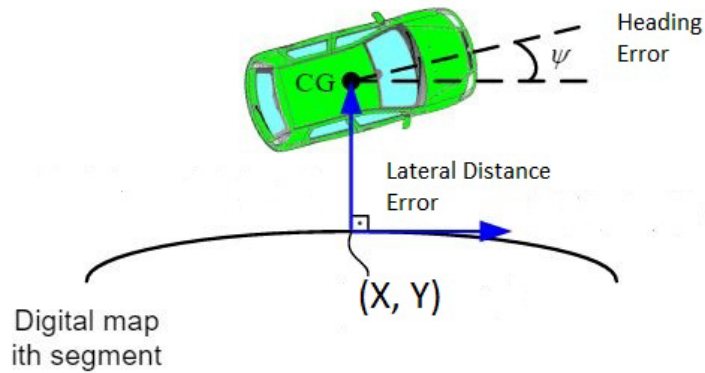

Figure 21. Illustration of error calculation.

## *Car Following*

### Cooperative Adaptive Cruise Control Design and Implementation

For the longitudinal control of the designed system, a speed controller is used when there is no traffic. In the case of a vehicle(s) in front, a Cooperative Adaptive Cruise Control (CACC) algorithm is designed. V2V communication is used if the vehicle in front has a modem and is capable of sending basic safety message set information and its acceleration. If this is not the case, the CACC algorithm automatically defaults to Adaptive Cruise Control (ACC) for car following. The developed CACC controls the inter-vehicle time gap between the target vehicle and ego vehicle using a feedforward PD controller. In this design, the feedforward information is the acceleration of the target vehicle which is communicated through a Dedicated Short-Range Communication (DSRC) modem. In this report, exemplary simulation and experimental results with the designed CACC system are presented. The presented results indicate that CACC improves the car following performance of the ego vehicle as compared to Adaptive Cruise Control alone.

The control structure of the designed CACC system is shown in the block diagram in Figure 22. String stability in both ACC and CACC modes are checked in advance. The designed control system is similar to the one designed and shown to be string -stable in [8]. Since the vehicle does not have built-in ACC, the low-level controller is designed as a gain-scheduled PI controller. As an upper controller, a PD controller with a feed-forward controller is used. To sustain the string stability a constant time headway spacing policy is employed [9]. The input of the feedforward controller is the acceleration of the target vehicle which is transmitted through DRSC radio communication.

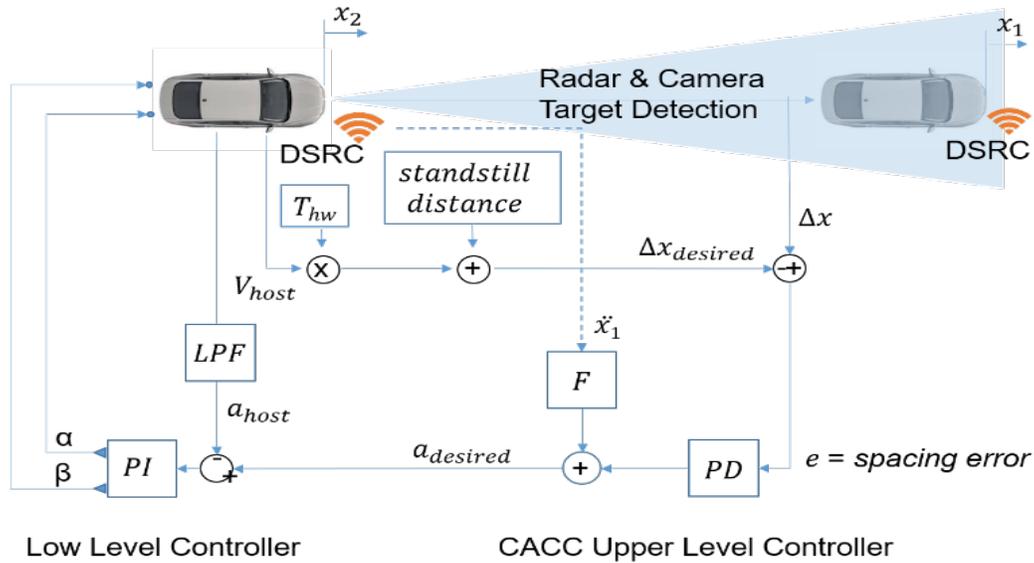

Figure 22. Cooperative Adaptive Cruise Controller (CACC) block diagram.

Development of the initial CACC model is done in the CarSim - MATLAB co-simulation environment [10]. CarSim is a vehicle simulation environment with the capability of simulating the dynamics of the vehicle. It can also simulate the target vehicle as a kinematic object. In the simulation, the target vehicle is driven by an Intelligent Driver model. By changing the desired speed and/or acceleration limits for the target vehicle, one can create different driving scenarios using the Intelligent Driver Model (IDM) [11]. Similarly, in the experiments, the target vehicle speed profile is chosen to be the same as the simulation target vehicle speed profile. The target vehicle speed profile is generated in real time with IDM driver similar to the simulation environment. In the result reported here, the target vehicle accelerates to 20 km/h and 25 km/h consecutively then it stops. In the CACC scenario, the simulated acceleration values for the target vehicle are broadcasted through DSRC OBU and are received by another OBU for the ego vehicle. One can see the experimental results for the ACC and CACC for 0.6 second time gap overlaid onto the simulation results on Figure 23.

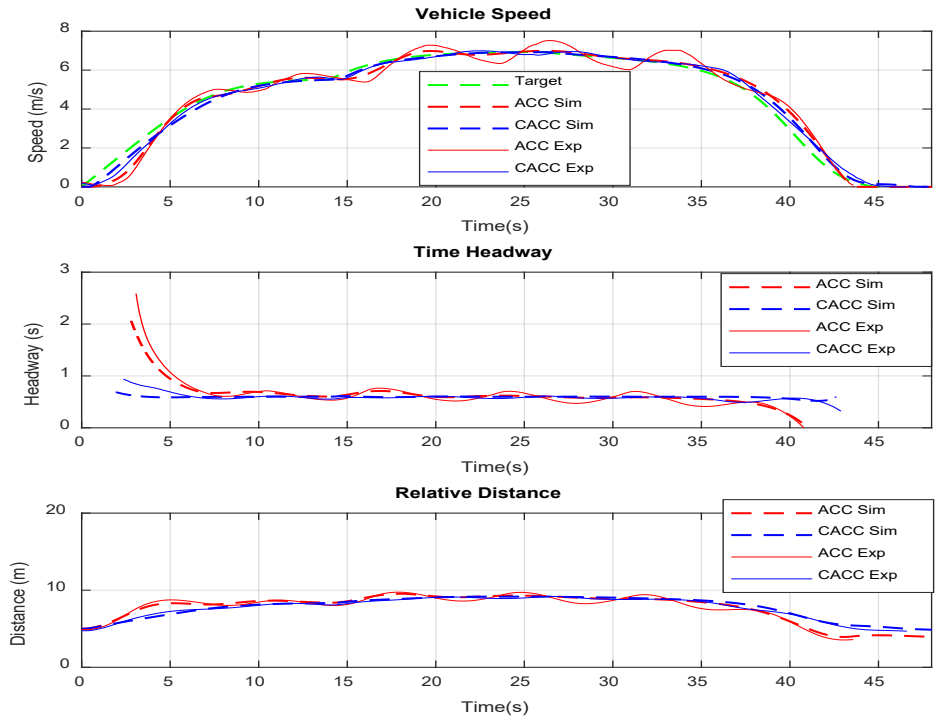

Figure 23. Comparison of ACC and CACC experimental results with simulation results for 0.6 second desired time gap.

The simulation results match with the experimental results. The small mismatches between the experiment and CarSim simulation are caused due to the fact that the CarSim vehicle model is not an exact model of the experimental vehicle. In response to the speed changes in the target vehicle speed profile, the CACC controlled vehicle starts accelerating and decelerating faster as compared to ACC. Thus, the CACC controller can follow the target vehicle more accurately. CACC time gap following performance is much better than the performance of ACC. More details on the designed system can be read in [10].

## *Pedestrian Collision Avoidance*

### Path Modification with Elastic Band

Pedestrian collision avoidance is introduced in this subsection but the method presented is applicable to collision avoidance with other obstacles also. After the position of the pedestrian is obtained, an alternative path to be followed is created if a collision is imminent. For our case, this path is created by modifying the existing points on the path according to the position of the pedestrian. Our path modification algorithm is based on the elastic band theory, where socially acceptable distance is also considered in modifying the deformed path. In elastic band theory, the initial path is deformed by internal and external forces acting on the band. Internal forces are spring forces which hold the band or the path together while external forces keep the band or path away from the pedestrian like artificial potential field forces. Figure 24 shows an initial path displayed as an elastic band which is deformed by both internal and external forces in the presence of a pedestrian.

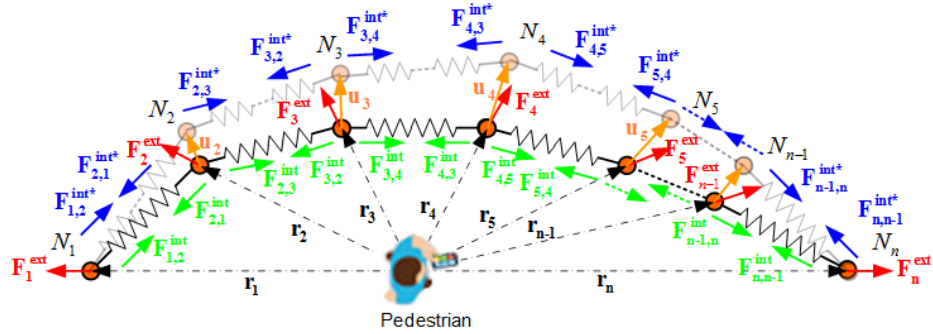

Figure 24. An initial path deformed by internal and external forces in the presence of a pedestrian (or other obstacle) on the path

The elastic band is a sequence of displaceable nodes denoted by $N_i$ in Figure 24 that initially correspond to the original local path of the autonomous shuttle in the vicinity of the detected pedestrian. The initial positions of the nodes $N_i$ with respect to the pedestrian are shown by position vectors $r_i$. Internal forces are formulated by adding springs with stiffness $k_s$ and spring force $\mathbf{F}_{i,j}^{int}$ acting on node $i$ due to the adjacent nodes $N_j$ with $j=i+1$ or $j=i-1$ for $i=1,2,…,n$. The function of internal forces is to hold the nodes or the local path together as a displaceable part of the route of the autonomous shuttle as shown in Figure 24. External forces $\mathbf{F}_i^{ext}$ acting on node $N_i$ with $i=1,2,…,n$ are defined once a pedestrian is detected to deform the band and hence the local path away from the pedestrian like artificial potential field forces. The external forces keep deforming the local path around the pedestrian who may be moving while the internal forces keep the nodes together in the form of a collision free path to be followed. $u_i$ for $i=1,2,…,n$ are the deformations of the nodes under the action of external and internal forces when a pedestrian is detected. The internal forces $\mathbf{F}_{i,j}^{int}$ become $\mathbf{F}_{i,j}^{int*}$ after the deformation of the local path. After a pedestrian is detected, external forces are applied and the nodes of the deformed path become the new positions $r_i+u_i$ for $i=1,2,…,n$ as determined by the balance of internal and external forces acting on the nodes.

The static balance of internal forces acting on node $N_i$ in Figure 24 before a pedestrian is detected are

$$\mathbf{F}_{i,i-1}^{int} + \mathbf{F}_{i,i+1}^{int} = k_s\left(\mathbf{r}_{i-1}-\mathbf{r}_i\right) + k_s\left(\mathbf{r}_{i+1}-\mathbf{r}_i\right) = 0 \qquad (7)$$

After the pedestrian is detected and external forces are applied, the static balance of internal and external forces acting on node $N_i$ in Figure 24 become

$$\mathbf{F}_{i,i-1}^{int*} + \mathbf{F}_{i,i+1}^{int*} + \mathbf{F}_i^{ext} = k_s \left( \mathbf{r}_{i-1} + \mathbf{u}_{i-1} - (\mathbf{r}_i + \mathbf{u}_i) \right) + k_s \left( \mathbf{r}_{i+1} + \mathbf{u}_{i+1} - (\mathbf{r}_i + \mathbf{u}_i) \right) + \mathbf{F}_i^{ext} = 0 \qquad (8)$$

which using the identity in Equation 7 becomes

$$\mathbf{F}_i^{ext} = -\left[ k_s \left( \mathbf{u}_{i-1} - \mathbf{u}_i \right) + k_s \left( \mathbf{u}_{i+1} - \mathbf{u}_i \right) \right] = -k_s \left( \mathbf{u}_{i-1} - 2\mathbf{u}_i + \mathbf{u}_{i+1} \right) \qquad (9)$$

The external force $\mathbf{F}_i^{ext}$ acting on node $N_i$ is calculated as a repulsive force using

$$\mathbf{F}_i^{ext} = \begin{cases} \left( \mathbf{F}_i^{ext} \right)_{max}, & |\mathbf{r}_i| < d \\ -k_e \left( |\mathbf{r}_i| - r_{max} \right) \dfrac{\mathbf{r}_i}{|\mathbf{r}_i|}, & d \leq |\mathbf{r}_i| \leq r_{max} \\ 0, & |\mathbf{r}_i| > r_{max} \end{cases} \qquad (10)$$

where |...| denotes the magnitude of the argument. $\left( \mathbf{F}_i^{ext} \right)_{max}$ in Equation (10) is used to saturate the repulsive force within $|\mathbf{r}_i| < d$ so that it does not go to infinity as $|\mathbf{r}_i| \to 0$. $k_e$ is the stiffness associated with the repulsive force and $r_{max}$ is the range of the repulsive force. Once a pedestrian is detected and localized with respect to the path of the vehicle using V2P communication, Equations 9 and 10 are solved to obtain the new coordinates $r_i + u_i$ for $i = 1, 2, \ldots, n$ of the locally deformed path. In the case of a moving pedestrian, the computations are repeated at each time step to continue to locally deform the obstacle avoidance path to be followed.

The distance $d$ in Equation 10 is also used to model the physical dimension of the autonomous vehicle $d_{vehicle}$. Noting that $|\mathbf{r}_i| < d$ is a circular region around the pedestrian to be avoided, $d$ is adjusted such the pedestrian stays within that region during any short duration relative displacement and the V2P communication delay of the car within the detection sampling instants. In the case of a moving pedestrian, the circular region $|\mathbf{r}_i| < d$ keeps moving with the pedestrian, requiring the local path modification calculations based on solving Equation 9 and 10 to take place within the steering control sampling time.

In socially acceptable collision avoidance, the circular region $|\mathbf{r}_i| < d$ is increased to also accommodate a pedestrian socially acceptable distance of about 1.5 m and $d$ is calculated using

$$d = d_{vehicle} + d_{pedestrians} + d_{social} \qquad (11)$$

where $d_{vehicle}$, $d_{pedestrian}$ and $d_{social}$ account for our the autonomous vehicle dimensions, the pedestrian possible motion between two V2P detection sampling instances and the social acceptance distance for pedestrians. A maximum possible pedestrian speed of 1.5 m/sec is used in computing $d_{pedestrian}$. While pedestrian (obstacle) detection was based on V2P communication here, the method is applicable for the case of detection using other sensors like lidar, camera and radar.

**Following the Modified Path**

A preview distance of 15 m to the V2P detected pedestrian is used in the HiL simulations and experiment here to modify the path based on the elastic band method of the previous sub-section. After modification of the points on the path with the elastic band method, the modified path should be followed instead of the actual path. Our past work involved fitting several segments of cubic polynomials to the deformed elastic band nodes. This method was implemented in real time for a stationary pedestrian in our previous work. However, this approach limited our real time computation speed for moving pedestrians where the curve fits had to be repeated at each time instant. For this reason, a simpler method for following the modified path with a point to point approach is formulated and used here. This method runs much faster since it uses a much faster computation to determine lateral deviation at each node. This method is illustrated in Figure 25.

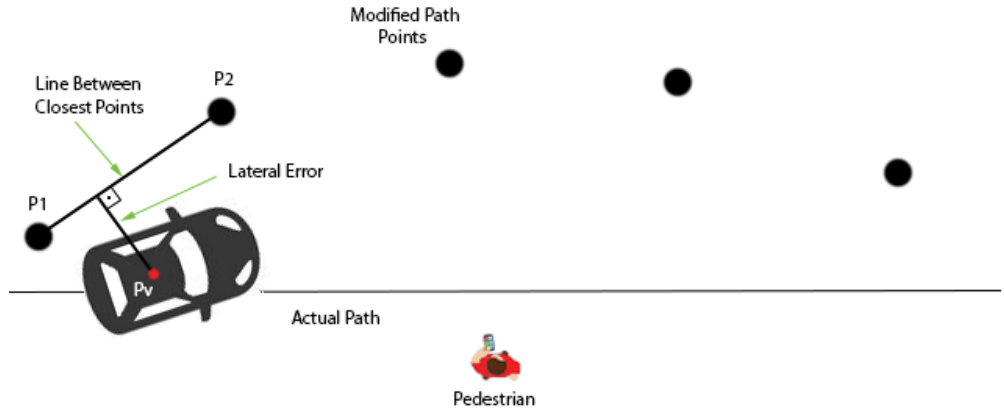

Figure 25. Lateral error calculation for pedestrian avoidance

In our new approach, the orthogonal distance to the line between two deformed nodes, which are the closest to the vehicle are calculated as shown in Figure 25. This distance is considered to be the lateral error for the modified path and used as the source of the lateral error for the steering controller, when there is a need to avoid the pedestrian. Calculation for a full elastic band method modified path consisting of 500 nodes and one pedestrian (obstacle) takes approximately 20 ms for one step while this time was approximately 200 ms for our previous method that used cubic curve fitting as explained. The lateral error $e_y$ for this method after the V2P based detection of a pedestrian on the path within the 15 m preview distance is

$$e_y = \frac{\det\left(\begin{bmatrix} a \\ b \end{bmatrix}\right)}{\|a\|} \tag{12}$$

$$a = P_2 - P_1 \tag{13}$$

$$b = P_v - P_1 \tag{14}$$

where *a* and *b* are row vectors calculated from Euclidean coordinates of closest points $P_1$, $P_2$ and vehicle's coordinate $P_v$. $\|a\|$ is the length of the vector *a*. This calculation is done at every step that needs obstacle avoidance behavior when there is a V2P detected pedestrian (obstacle) who is nearby. A flowchart that presents the steps of the overall algorithm and decision making is shown in Figure 26.

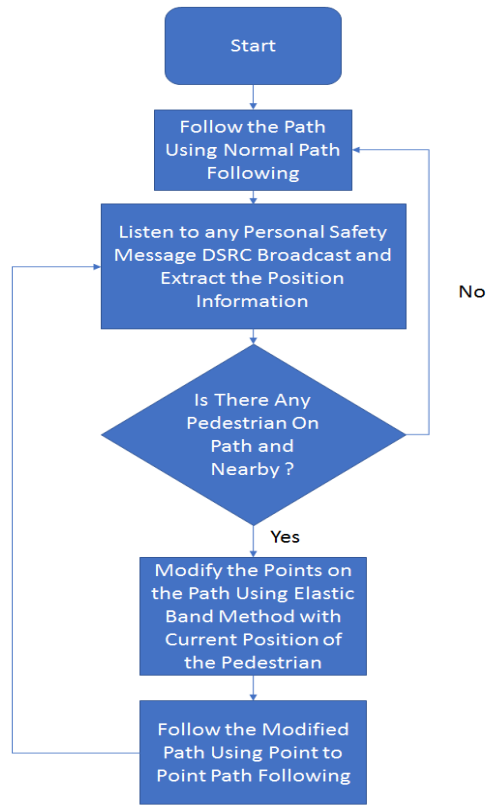

Figure 26. Flowchart of the algorithm

## *Processing/Perception*

### Nvidia Drive PX2

The Nvidia Drive PX2 AutoChauffeur platform features two discrete GPUs and various functions for the development of autonomous vehicles. The Drive PX2 is intended to be used for online processing of trained neural networks to enable Autopilot and self-driving functionalities. The platform features already trained networks like DriveNet, LaneNet, and OpenRoadNet. Some functions that can be easily implemented with the help of these networks include basic object detection and tracking, free-space detection, and lane detection. Such algorithms were run against the data we collected in the Easton Town Center shopping area. The results give promise to the development of autonomous vehicles using the Drive PX2 as the main controller in the vehicle. The Drive PX2 neural networks are based on the use of camera sensors and are suitable for autonomous shuttles in dry weather conditions.

The lane-detection algorithm provides an indicator on the lane the vehicle is currently on along with markings on the adjacent lanes when they are present. The algorithm uses Nvidia's pre-trained neural network LaneNet as shown in Figure 27.

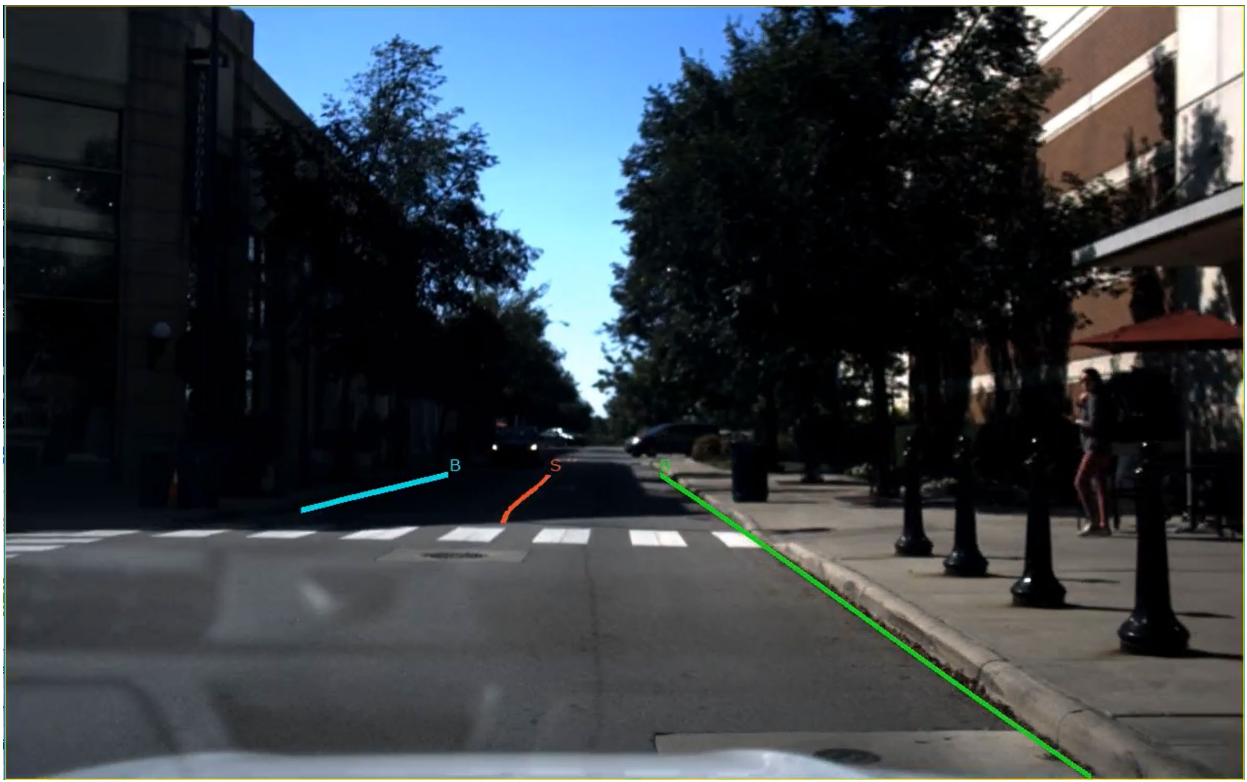

Figure 27. Lane detection algorithm.

The free-space detection algorithm provides drive-able collision-free space and labels each pixel on the boundary with colors indicating whether the pixel is associated with a vehicle, pedestrian, curb or other. The algorithm uses Nvidia's already pre-trained neural network OpenRoadNet as shown in Figure 28.

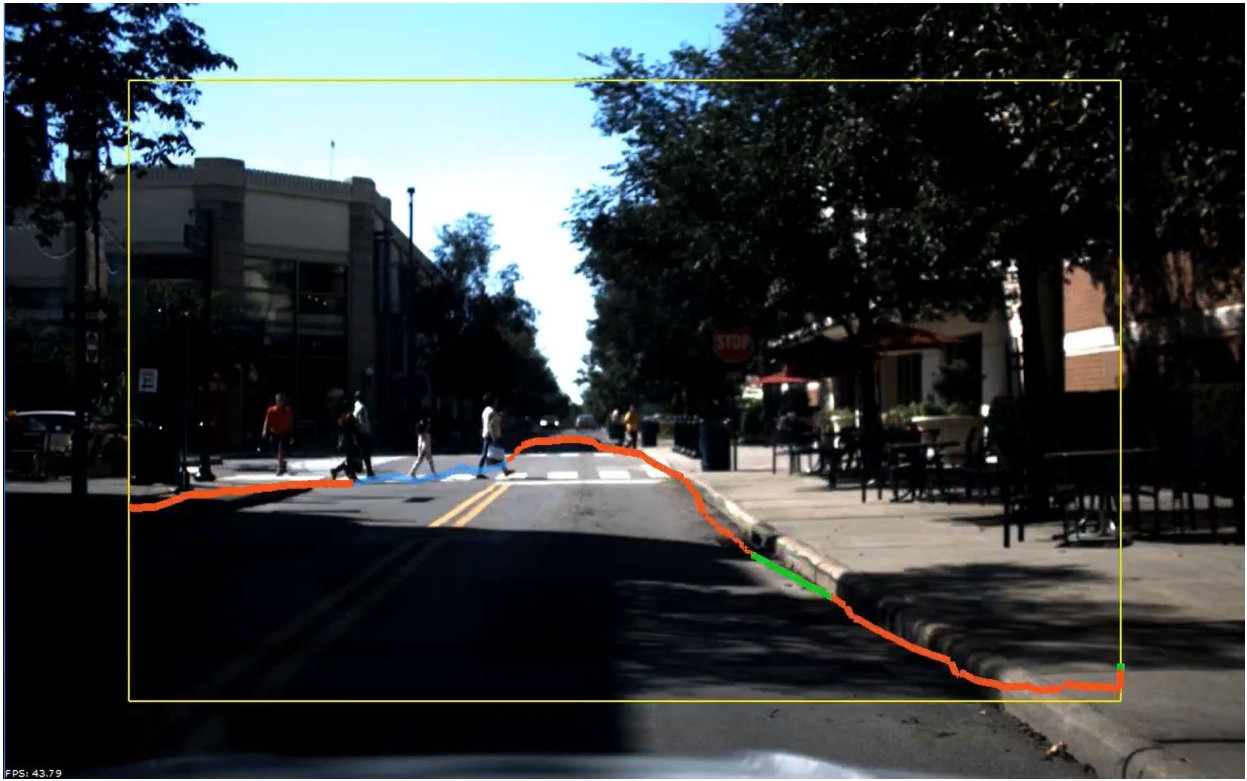

Figure 28. Free space algorithm.

Furthermore, the Nvidia PX2 architecture provides support for other external sensors which can be utilized in conjunction with some of the more sophisticated functions in the library to detect road parameters and actuate the vehicle via the car controller box.

## Detection and Classification with YOLO

For the perception, one of the popular deep neural network methods YOLO [12] has been implemented in our experimental vehicle to detect the relevant objects on the road. In our case, relevant information can be considered as vehicles, pedestrians, cyclists, traffic lights stop signs and so on. For this perception task, a camera mounted at the windshield of our experimental vehicle is connected to the in-vehicle computer which is equipped with an Nvidia GPU. Example detection results for the deployed algorithm can be seen in Figure 29.

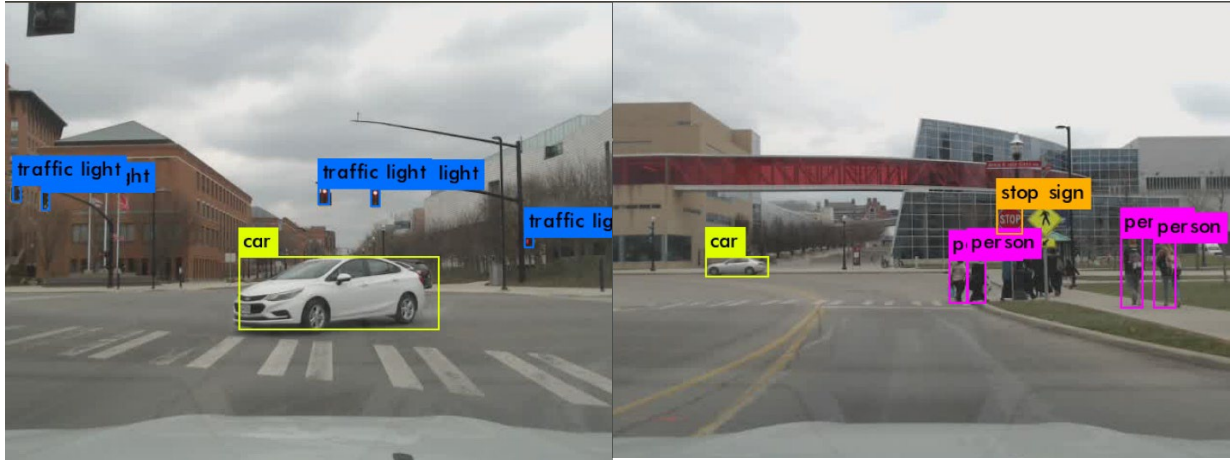

Figure 29. Example perception results with YOLO real-time object detection algorithm at the Ohio State University main campus.

## Semantic Segmentation

We investigate several neural network architectures that are tailored to the semantic segmentation task for outdoor scenarios.

*FC-DenseNet*

FC-DenseNet [13] is adopted from DenseNet architecture [14], which is built from dense blocks and pooling layers. Each dense block is iteratively concatenated with previous feature maps. Jégou et al. [13] suggest to extend DenseNet to a fully connected network similar to fully connected ResNet [15]. Regular fully connected networks use convolution, upsampling operations, and skip connections. In order to make DenseNet fully connected, the authors use a dense block rather than a convolution operation. Figure 30 illustrates an overview of FC-DenseNet.

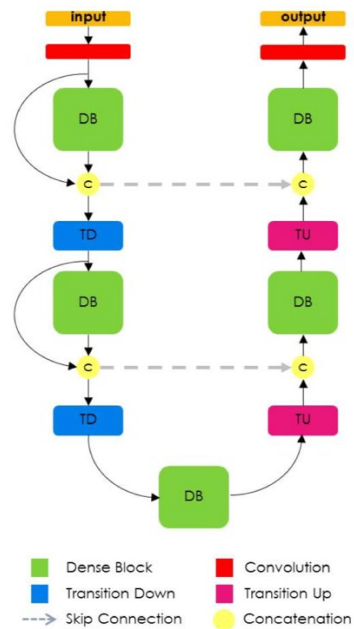

Figure 30. FC-DenseNet architecture. Transition Up (TU) is the upsampling operation, which is a 3x3 transposed convolution. Transition Down (TD) is consist of batch normalization, ReLU, 1x1 convolution, dropout with p = 0.2, and 2x2 max pooling [13].

*BiSeNet*

The second model is BiSeNet [16], which is designed to perform real-time semantic segmentation. The two main components of BiSeNet are the Spatial path and the Context path. Spatial path tries to preserve the input resolution by encoding rich spatial information using large feature maps. The Spatial path is comprised of three convolutional layers, each followed by batch normalization [17] and ReLU [18]. The Context path is designed to provide large receptive fields. The Context path is a lightweight backbone architecture such as Xception [19] that downsamples feature maps in order to obtain high level context information.

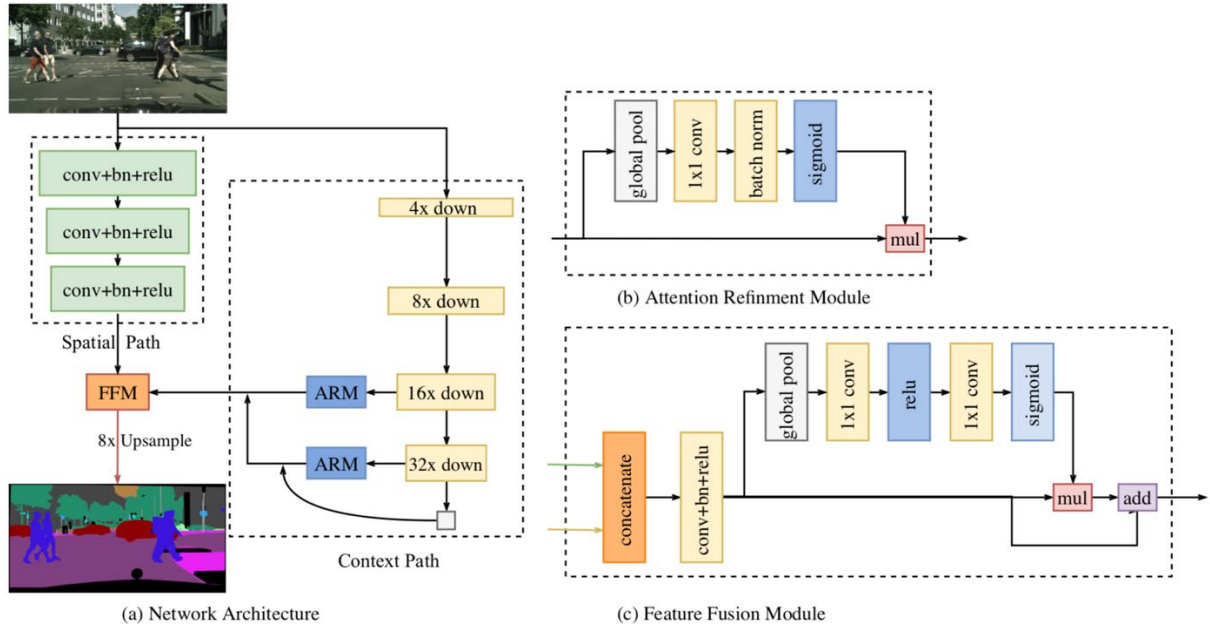

Figure 31. BiSeNet architecture. a) The main architecture with Spatial path and Conext path. b) Attention Refinement Module (ARM). c) Feature Fusion Module that is responsible for combining feature maps of different information level [16].

The Context path also includes an Attention Refinement Module (ARM), which is a global pooling average followed by 1x1 convolution, batch normalization and a sigmoid layer. The final output of this module is the multiplication of the input of this module by the output of the sigmoid layer. ARM is designed to capture global context information and computes a vector that guides the feature learning process. Since the Context path and the Spatial path contain different information about the scene, one cannot simply add feature maps resulting from these components. The Spatial path provides features that are rich in detail, while the Context path generates features that are high level. Combing these features is done via a module called Feature Fusion Module (FFM). FFM concatenates features, applies batch normalization, then uses a global average pooling. Figure 31 illustrates the BiSeNet architecture and its modules.

**Dense-ASPP**

A major problem in autonomous driving is the significant change in objects scales. Dense-ASPP [20] tackles this problem by using Atrous Spatial Pyramid Pooling (ASPP). ASPP [21] offers large receptive fields while preserving the spatial resolution. ASPP concatenates multiple atrous-convolved features with different dilation rates to generate multi-scale features. In order to make features dense enough for autonomous driving scenarios, ASPPs are densely connected. Dense-ASPP is basically a set of densely connected atrous convolutional layers (Figure 32). For a more detailed description of this network please refer to [20].

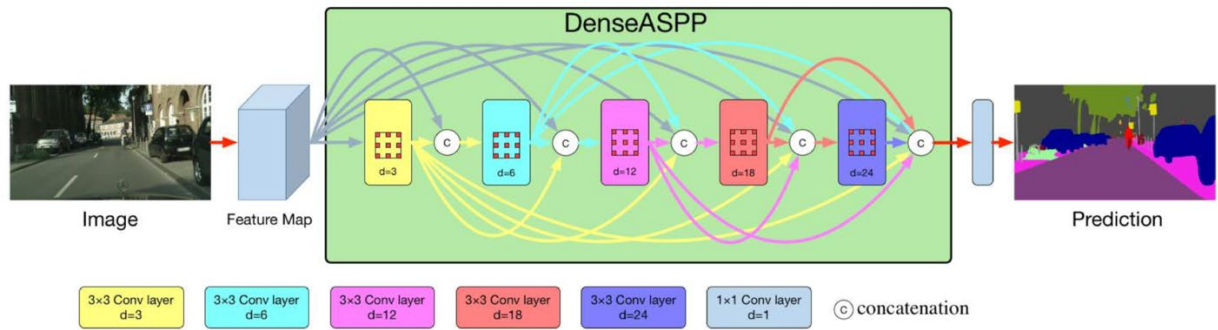

Figure 32. DenseASPP architecture [21].

**Semantic Segmentation Experiments**

We investigate the performance of the three models: BiSeNet, DenseASPP, and FC-Dense on the CamVid dataset. CamVid is an outdoor scene dataset which is recorded at 15 fps from the driver's point of view. It contains 367 training, 101 validation, and 233 test images totaling up to 701 images. The dimension of images is 960x720, and the dataset includes 32 semantic categories: Animal, Pedestrian, Child, Rolling cart/luggage/pram, Bicyclist, Motorcycle/scooter, Car (sedan/wagon), SUV / pickup truck, Truck / bus, Train, Misc., Road, Shoulder, Lane markings drivable, Non-Drivable, Sky, Tunnel, Archway, Building, Wall, Tree, Vegetation misc., Fence, Sidewalk, Parking block, Column/pole, Traffic cone, Bridge, Sign / symbol, Misc. text, Traffic light, Other.

Some categories comprise less than 0.1% of the dataset. Therefore, we do not report them in the following tables.

Table 2. Visual comparison between BiSeNet, FC-DenseNet, DenseASPP.

| Image | BiSeNet | DenseASPP | FC-DenseNet | Ground Truth |
|---|---|---|---|---|

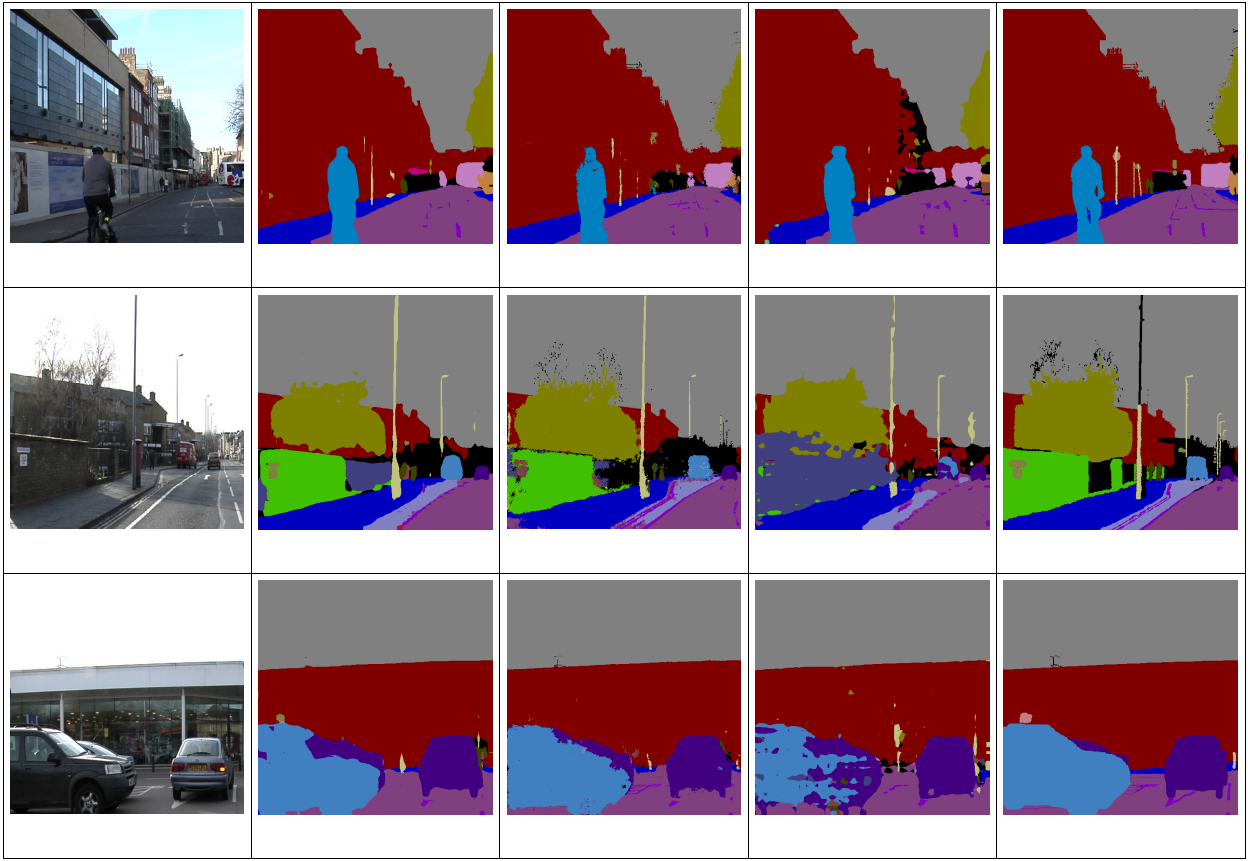

Table 3 Accuracy, precision, recall, F1-score, and mean IoU for BiSeNet, DenseASPP, and FC-DenseNet. Values are given in percentage (%).

|  | Accuracy | Precision | Recall | F1-score | Mean IoU |
|---|---|---|---|---|---|
| **BiSeNet** | 84.67 | 86.07 | 84.67 | 84.21 | 49.72 |
| **DenseASPP** | 77.03 | 78.38 | 77.03 | 75.35 | 39.12 |
| **FC-DenseNet** | 85.96 | 88.11 | 85.96 | 86.0 | 49.57 |

Table 4 Class-wise accuracy for BiSeNet, DenseASPP, and FC-DenseNet. Values are in percentage (%).

|  | Bicyclist | Building | Car | Fence | Pedestrian | Sidewalk | Sign | Sky | Tree | Road | Pole |
|---|---|---|---|---|---|---|---|---|---|---|---|
| **BiSeNet** | 88.30 | 85.88 | 82.50 | 84.02 | 67.83 | 82.43 | 75.02 | 95.14 | 84.46 | 91.75 | 51.99 |
| **DenseASPP** | 83.64 | 77.89 | 82.56 | 85.34 | 48.91 | 61.70 | 63.77 | 87.00 | 80.65 | 82.80 | 38.38 |
| **FC-DenseNet** | 88.18 | 86.50 | 79.57 | 81.85 | 58.90 | 84.41 | 61.87 | 96.12 | 86.42 | 91.58 | 38.90 |

|  | Cart | Child | Lane Markings | Motorcyclist | Parking | Traffic Light | Wall |
|---|---|---|---|---|---|---|---|
| **BiSeNet** | 73.38 | 96.11 | 39.20 | 97.61 | 87.93 | 79.36 | 74.48 |
| **DenseASPP** | 70.59 | 93.79 | 32.83 | 97.64 | 90.17 | 71.73 | 61.22 |
| **FC-DenseNet** | 70.47 | 95.44 | 50.84 | 97.62 | 89.99 | 74.99 | 73.12 |

|  | Road shoulder | SUV | Bus | Other Moving Obj. | Vegetation | Void |
|---|---|---|---|---|---|---|
| **BiSeNet** | 92.36 | 79.21 | 94.26 | 87.64 | 85.39 | 42.30 |
| DenseASPP | 92.67 | 67.39 | 92.23 | 84.07 | 80.73 | 41.52 |
| FC-DenseNet | 95.08 | 78.43 | 92.78 | 85.40 | 87.15 | 47.00 |

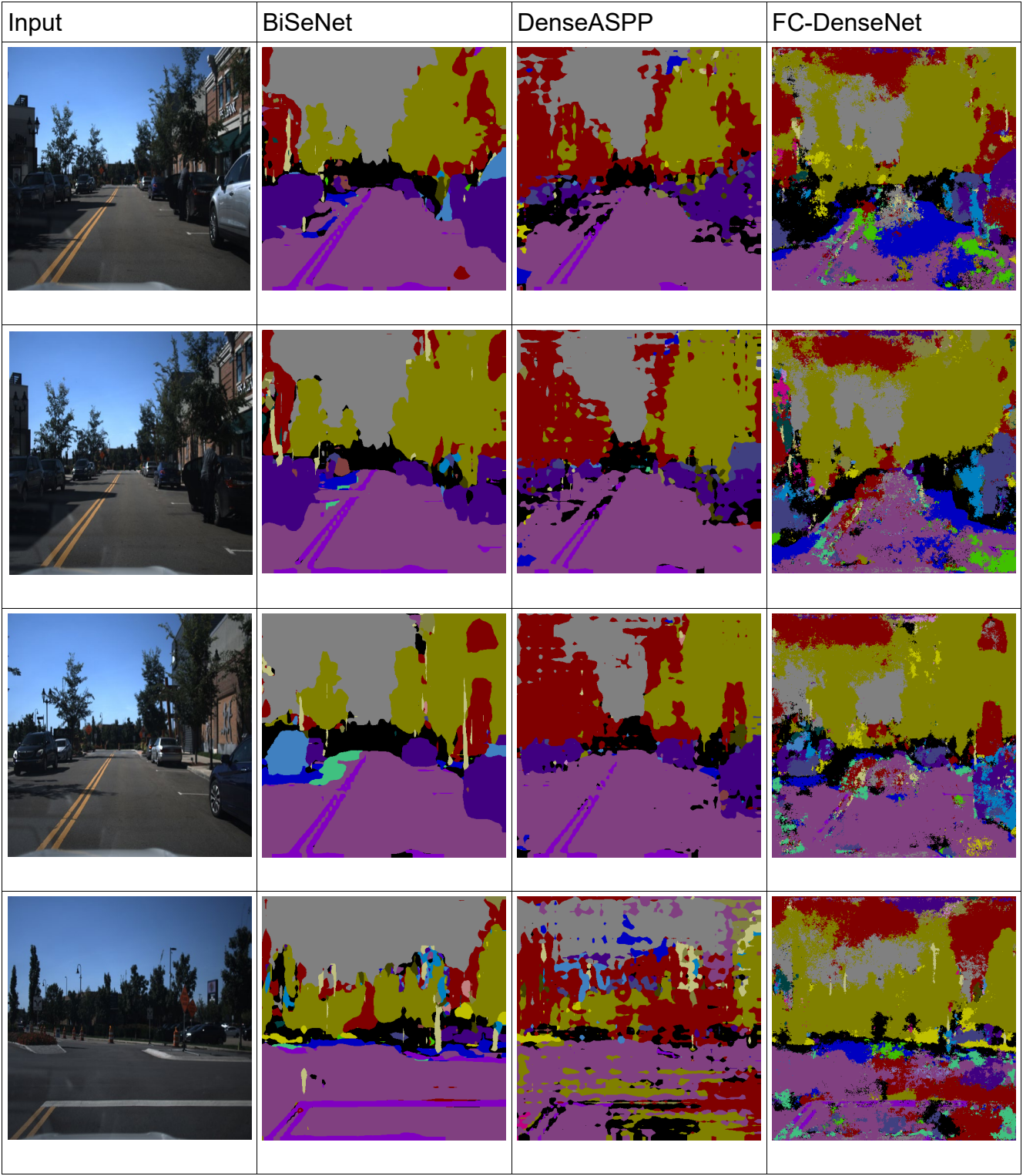

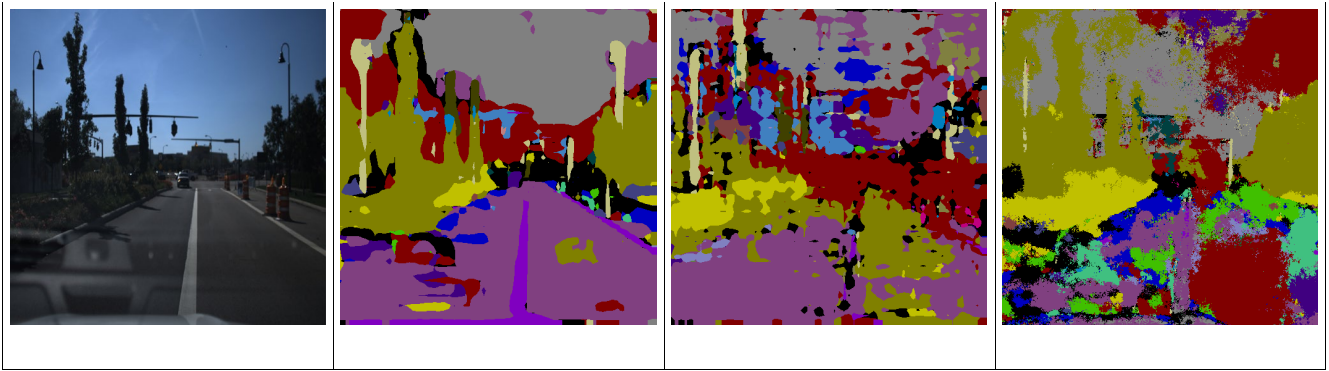

### *Decision Making*

At the core of autonomous vehicle control system, the supervisory controller plays a very important role in determining the appropriate state and subsequent course of action when faced with different situations. Decision making has been a research topic that has drawn broad attention and great progress is being made with the various methods that are deployed.

To take care of the uncertainties in the environment, many research efforts in decision making have utilized the probabilistic method for errors that occur in the perception part of autonomous vehicle. Markov Decision Process is one of the most popular methods. A lot of work has been done in decision making based on Markov Decision Process or POMDP to deal with the uncertainties existing in perception and situational awareness. Also, since artificial intelligence is also commonly used and well developed, decision making based on machine learning has also been studied and implemented widely in autonomous driving.

Due to the heavy computation and uncertainties in the probabilistic method and machine learning method, the platform for supporting decision making has to have a powerful computer like the NVIDIA GPUs used in our hardware architecture for good performance. The rule-based decision-making framework like the simple example in Figure 33 is used in this report because of lower computation complexity and ease of implementation on most of the vehicle platforms we use. Work on replicable and scalable probabilistic or machine learning based decision making is work in progress.

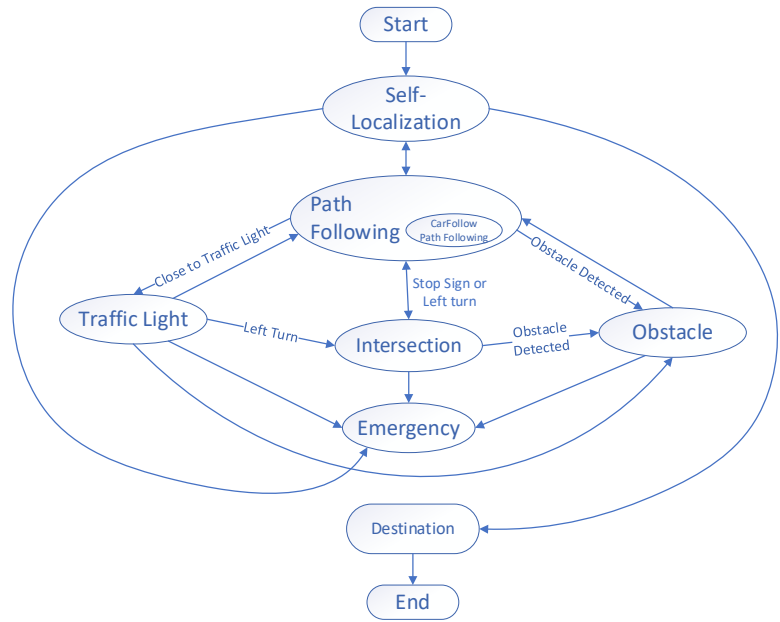

Figure 33. Diagram of Framework for Decision Making in the Unified Autonomous Vehicle Architecture with some of switching conditions shown.

## Evaluation - Data Collection and Sharing

We used the Ford Fusion sedan to collect multiple sensor raw data for use in autonomous driving research. The data collection has resulted in eight datasets which we share in a box site and provide interested researchers with a publicly accessible link. We will continue these data collection and sharing exercises in more of the planned autonomous shuttle routes in Columbus, Ohio. The data collection covers areas including downtown COSI routes (Figures 34-36), Easton town center (Figure 37), road connecting CAR with CAR West (Figure 38), the parking lot around CAR West (Figure 39) center and some highway routes around Columbus (no figure display). The data is classified based on the area and sensors used for data collection. Further development will include data from other areas and data labelling for various ways of utilization.

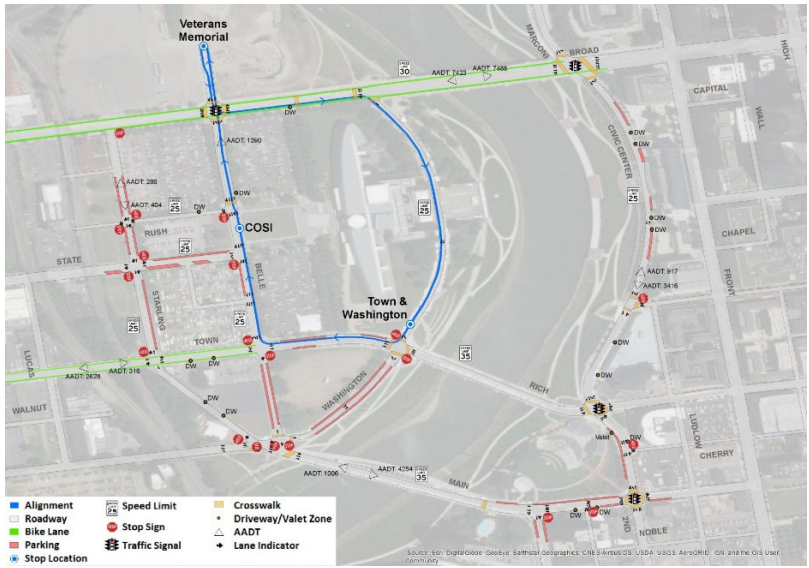

Figure 34. COSI Loop 1.

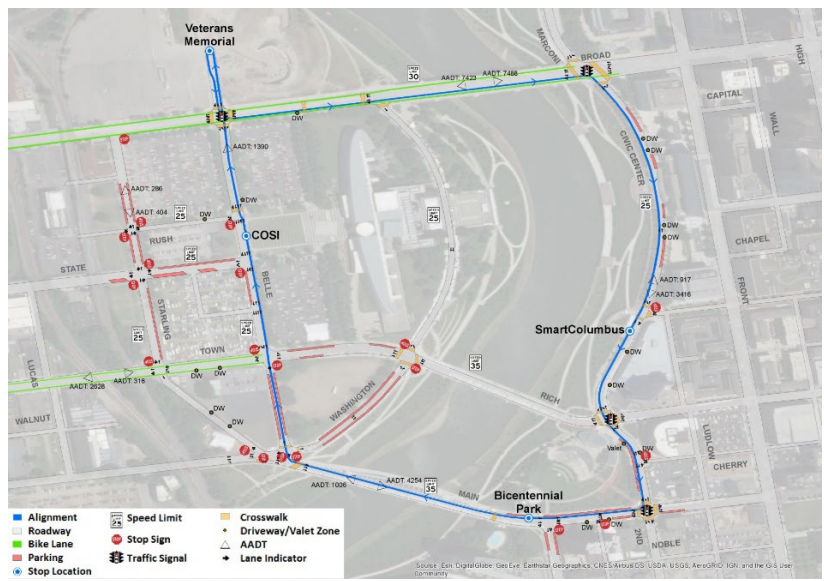

Figure 35. COSI Loop 2.

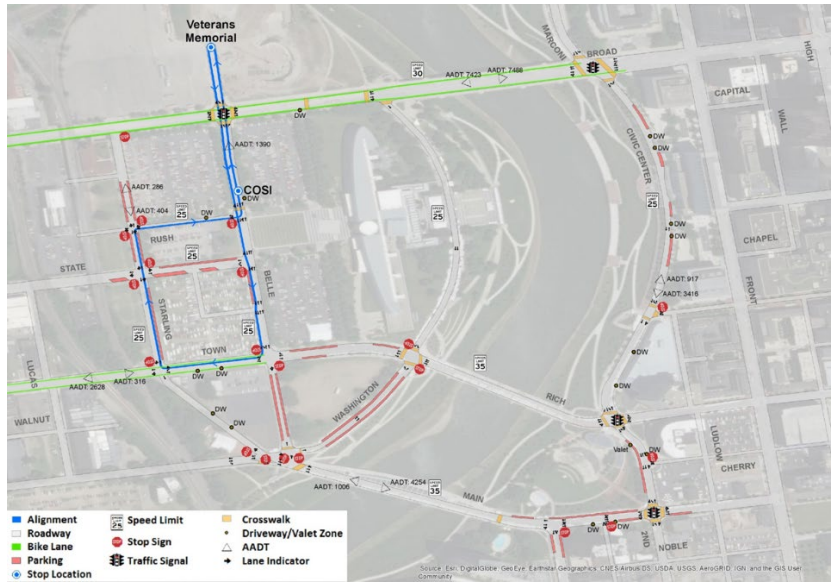

Figure 36. COSI Loop 3.

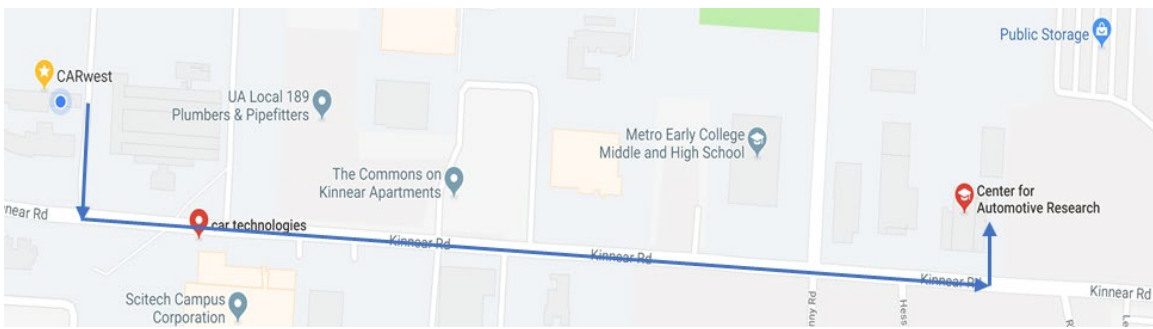

Figure 37. CAR to CAR West route.

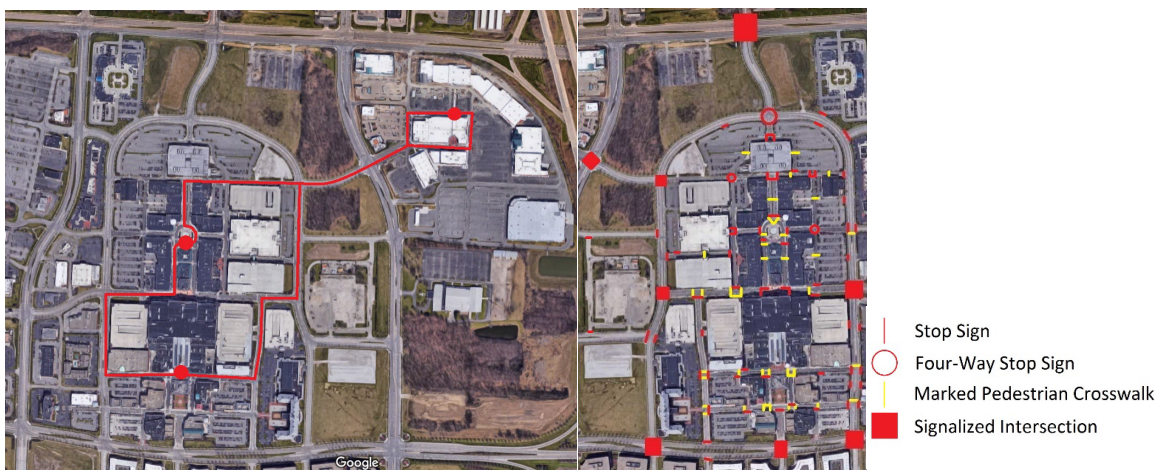

Figure 38. Easton Town Center with position of road signs.

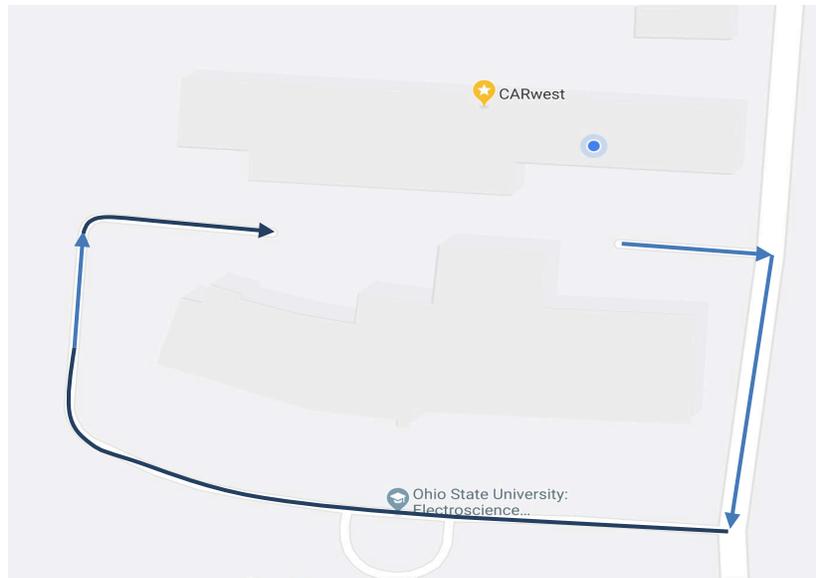

Figure 39. Parking lot at CAR West.

The sensor suite used includes Velodyne VLP-16 16 channel 3D Lidar, Ouster OS1-64 64 channel 3D Lidar, Pointgrey Camera with 30 fps and OXTS xNav 550 GPS/IMU data. From those sensors, different types of data are present in the dataset, which are, pointcloud data, images, GPS location data and IMU data. All the data are raw data stored in the format of a ROS bag file in order to make the data accessible to a wider audience. The rosbag play command can be used to playback the data. For the convenience of storing and using, we split the recording to consecutive files of 20 seconds each and used a sequential ordering starting from 0, meaning the first 20 seconds of data. The data is available for download at: https://osu.box.com/s/wl3ax8iywcciz0swzikdfbo7mzamrhey.

## *Evaluation - HiL Simulator*

To extensively evaluate the performance of the developed low level controllers, along with both high level supervisory control, decision making and sensor placement, a hardware-in-the-loop simulator is employed. The HIL simulator is a platform where we can run several test scenarios in a realistic environment in real time. It also allows us to test our actual experimental hardware in the system. In the simulations, a high fidelity CarSim vehicle model is used to simulate the lateral and longitudinal motions of the vehicle alongside the soft perception sensors. While this model is running in the dSpace Scalexio HIL platform, a MicroAutobox controller, an in-vehicle PC with Nvidia GPU and DSRC modems emulate the actual control and communication scheme in this simulation environment. Control strategy and decision making were implemented inside the MicroAutobox while high fidelity vehicle dynamics model runs in real time on SCALEXIO. Traffic is added using other vehicles in Carsim and co-simulation with SUMO and PTV Vissim. The HIL simulator setup can be seen in Figure 40.

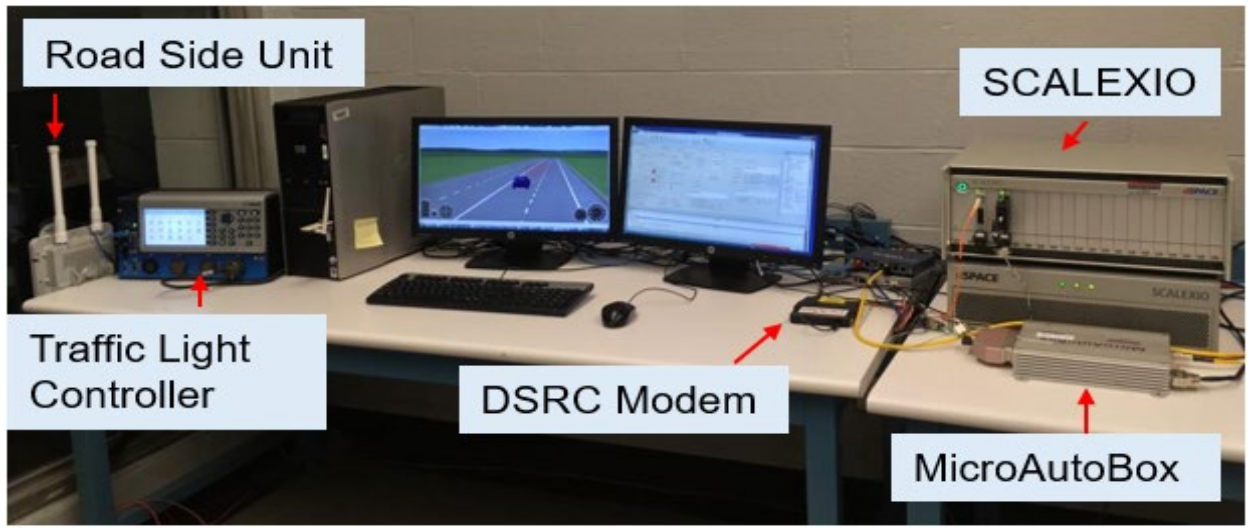

Figure 40. Hardware-in-the-Loop simulator setup.

The OSU AV pilot test route from CAR West (our lab location) to CAR (Center for Automotive Research – our main research center) shown in Figure 37 was chosen and constructed in CarSim to evaluate the vehicle's decision making and low level control performance. To incorporate the real traffic into simulation, information about other vehicles on the road were imported from SUMO software. Placement and field of view (FOV) of the sensors implemented as soft sensors in the HIL simulation can be seen in Figure 41.

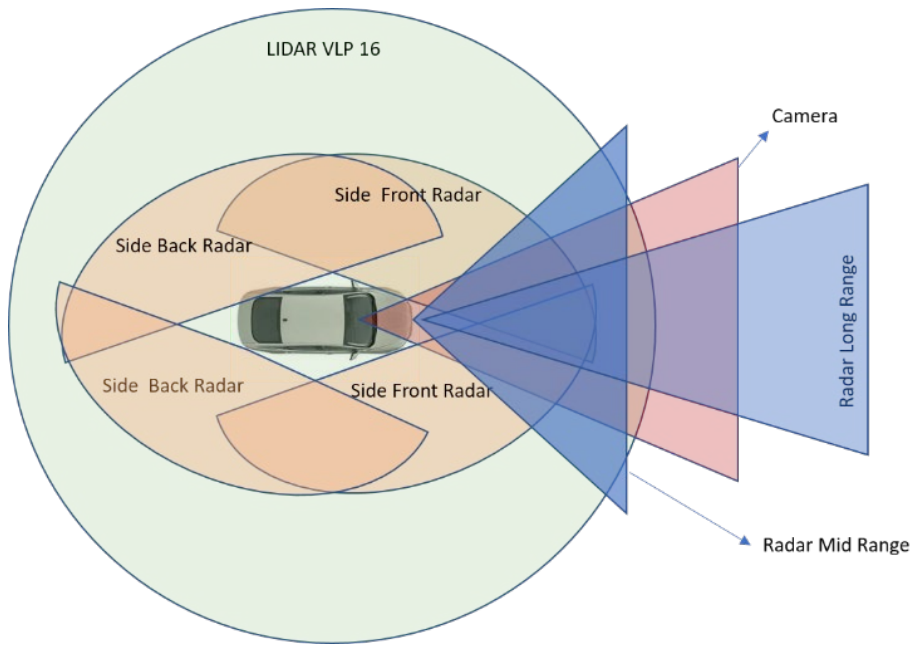

Figure 41. Placement and FOV of the sensors on the car.

Scenarios including stop sign, crossing traffic and traffic lights were simulated on the route in Figure 37 to test the decision-making strategies introduced in previous sections. Figure 42 shows the vehicle speed, steering and tracking error along the route. Since sharp turns appear when entering and exiting the main straight road, lateral error *e* was expectedly high for these two cases, but the look-ahead error *y* which combines the lateral and heading angle error was still relatively small. A video animation of this soft HIL AV test is present at: https://youtu.be/_yWiZWP0Rag.

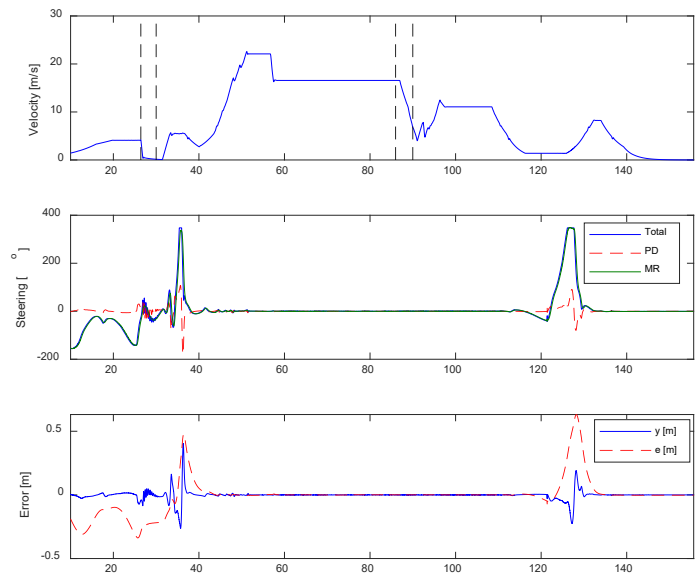

Figure 42. CARWest-to-CAR AV pilot test route HIL simulation results.

## *Evaluation – Scalability of Path Tracking Experiments*

After the controller design, real world experiments were done for the purpose of analyzing the effectiveness of both controller parameters and overall unified approach. The path following algorithm developed for the sedan is scaled to the small electric vehicle and tested as it is driven autonomously on an open field. An oval path is generated for the vehicle to follow and loop two times in order to test both performance and repeatability. First path following experiment is done by using the lateral controller coefficients designed for the sedan vehicle without any change and it works except for the relatively large path tracking error as low speed is used. The second experiment is done with the controller coefficients re-tuned for the small electric vehicle. Results for both of the experiments are shown on the same plots in Figures 43 and 44 to make comparison easier.

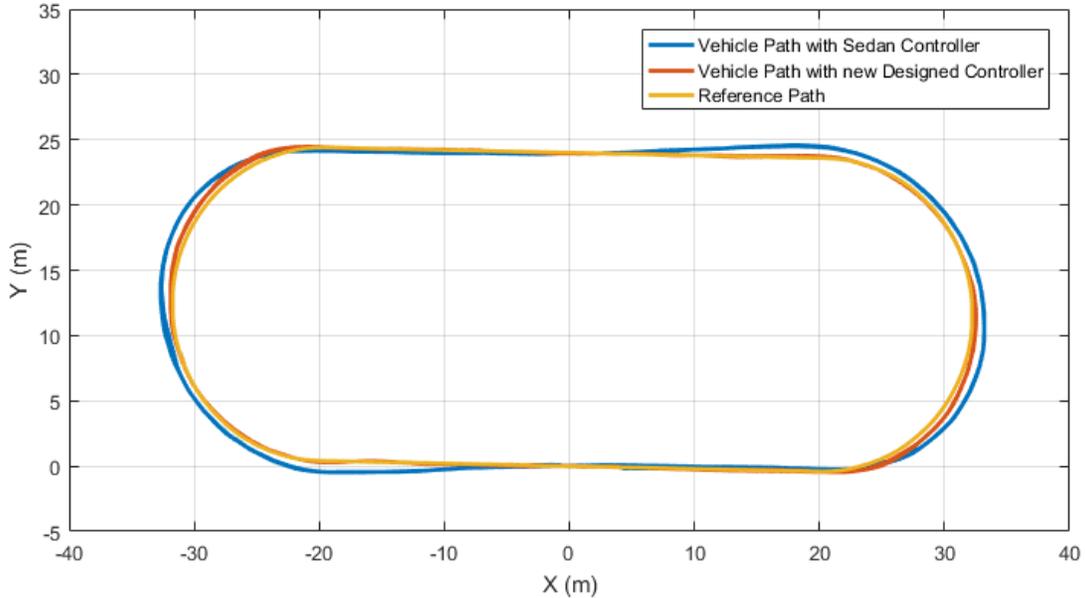

Figure 43. Path following comparison of Dash EV for two different controllers.

In Figure 43, we can see the reference path with comparison to the vehicle path while it is doing autonomous path following. We can also see two different controllers and how the performance is affected by changing coefficients according to the system parameters scaling down from one vehicle to other one. Sedan controller is not able to keep up with path when there is a curve and creates a significant error. There is also a very small error while it is following a straight road. Dash re-tuned controller on the other hand, is better in both of following a straight line or following a curved road. Moreover, the vehicle follows the same path for both its first and second lap in the experiments with both of the controllers, meaning repeatability is good for both of the controllers.

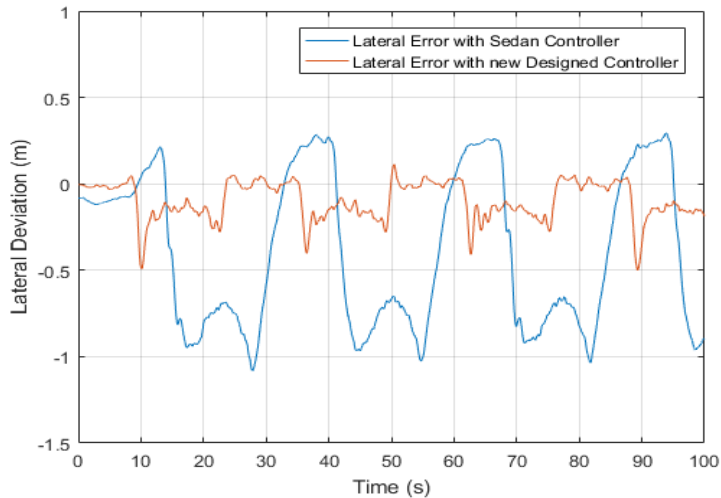

Figure 44. Lateral error comparison for two different controllers.

In Figure 44, we can see the lateral deviations in the course of approximately two laps of the path for both of the controllers. Lateral error from the experiment with the sedan controller yields higher peak values and has RMS value of 0.5636 whereas the other controller yields much lower peak values and has an RMS error value of 0.1443, which is very close to the value of 0.121. This is similar performance to that of the sedan vehicle with the sedan controller. It should be noted that the re-tuning of the control gains was conducted using a robust parameter space design and took a very short time to formulate and execute. A video of the Dash EV vehicle during this path tacking run is available at: https://youtu.be/WQjOZoPdQh4. The collision avoidance maneuvering application was also scaled successfully from the sedan vehicle to the Dash EV vehicle.

### *Evaluation – Proof of Concept Testing On-Demand Shuttle*

A limited scale proof-of-concept demonstration of on-demand AV shuttle use as a first mile / last mile solution was conducted during the first year of the project in the parking lot around our lab. The video of this demonstration is available at: https://youtu.be/K9dCd4ofYxA.

### **Publications**

The SmartShuttle sub-project resulted in the following publications that acknowledge project support. Copies of the papers (published ones) in this list are appended to the end of this report.

Journal Papers

1. Gelbal, S.Y., Aksun-Guvenc, B., Güvenç, L., "Elastic Band Collision Avoidance of Low Speed Autonomous Shuttles with Pedestrians," International Journal of Automotive Technology, under review.

2. Zhu, S., Gelbal, S.Y., Aksun-Guvenc, B., Güvenç, L., "Parameter-space Based Robust Gain-scheduling Design of Automated Vehicle Lateral Control," IEEE Transactions on Vehicular Technology, *under review*.

3. Bowen, W., Gelbal, S.Y., Aksun-Güvenç, L., Güvenç, L., 2018, "Localization and Perception for Control and Decision Making of a Low Speed Autonomous Shuttle in a Campus Pilot Deployment," SAE International Journal of Connected and Automated Vehicles, paper number 12-01-02-0003, pp. 53-66.

Conference Papers

1. Cantas, M.R., Gelbal, S.Y., Aksun-Güvenç, B., Güvenç, L., 2019, "Cooperative Adaptive Cruise Control Design and Implementation," WCX19: SAE World Congress Experience, April 9-11, Detroit, Michigan, Session AE 504 Intelligent Transportation Systems, SAE Paper Number 2019-01-0496.

2. Wang, H., Güvenç, L., 2019, "Discrete-Time Robust PD controlled system with DOB/CDOB compensation for high speed autonomous vehicle path following," WCX19: SAE World Congress Experience, April 9-11, Detroit, Michigan, Session AE 501 Intelligent Vehicle Initiative, SAE Paper Number 2019-01-674.

3. Li, Xinchen, Zhu, S., Gelbal, S.Y., Cantas, M.R., Aksun-Güvenç, B., Güvenç, L., 2019, "A Unified, Scalable and Replicable Approach to Development,

## Conclusions and Recommendations

The final project report for the SmartShuttle sub-project of the Ohio State University has been presented in this report. SmartShuttle was a two year project where the unified, scalable and replicable automated driving architecture introduced by the Automated Driving Lab of the Ohio State University was further developed, replicated in different vehicles and scaled between different vehicle sizes. The project approach, some of the results obtained and links to some videos and raw datasets were presented in the report. The readers are referred to our publications for more detailed information.

Addendum: The publications resulting from the work reported here and listed above have been added to the list of references as [22] to [32] except for those that already existed in the original references list. Reference 10 in the original report was the same as reference 1 by mistake. This was corrected. The following relevant references (some of them being more recent) were added to the list of references [33] to [53].

using a hardware in the loop steering test rig," IEEE Intelligent Vehicles Symposium, pp. 852-859.